\def\year{2021}\relax
\documentclass[letterpaper]{article} 
\usepackage{aaai21}  
\usepackage{times}  
\usepackage{helvet} 
\usepackage{courier}  
\usepackage[hyphens]{url}  
\usepackage{graphicx} 
\usepackage{natbib}  
\usepackage{caption} 
\usepackage{hyperref} 
\usepackage{amsmath}
\usepackage{times}
\usepackage{helvet}
\usepackage{courier}
\usepackage{caption}
\usepackage{subfigure}
\usepackage{amsthm,amsmath,amssymb}
\usepackage{mathrsfs}
\usepackage{bbm}
\usepackage[noend]{algpseudocode}
\usepackage{algorithmicx,algorithm}

\title{BGC: Multi-Agent Group Belief with Graph Clustering}

\usepackage{background}
\backgroundsetup{contents={%
    \begin{tikzpicture}[overlay,font=\sffamily]
    \end{tikzpicture}
  },color=blue,scale=1,angle=0}

\begin{document}

\author {
    Tianze Zhou,\textsuperscript{\rm 1}
    Fubiao Zhang, \textsuperscript{\rm 1}
    Pan Tang \textsuperscript{\rm 1} 
    Chenfei Wang \textsuperscript{\rm 2} \\
}
\affiliations {
    \textsuperscript{\rm 1} Beijing Institute of Technology\\
    \textsuperscript{\rm 2} Boston University \\
    tianzezhou@bit.edu.cn, wang1029@bu.edu.cn
}

\maketitle

\begin{abstract}
Recent advances have witnessed that value decomposed-based multi-agent reinforcement learning methods make an efficient performance in coordination tasks. Most current methods assume that agents can make communication to assist decisions, which is impractical in some situations. In this paper, we propose a semi-communication method to enable agents can exchange information without communication. Specifically, we introduce a group concept to help agents learning a belief which is a type of consensus. With this consensus, adjacent agents tend to accomplish similar sub-tasks to achieve cooperation. We design a novel agent structure named Belief in Graph Clustering(BGC), composed of an agent characteristic module, a belief module, and a fusion module. To represent each agent characteristic, we use an MLP-based characteristic module to generate agent unique features. Inspired by the neighborhood cognitive consistency, we propose a group-based module to divide adjacent agents into a small group and minimize in-group agents' beliefs to accomplish similar sub-tasks. Finally, we use a hyper-network to merge these features and produce agent actions. To overcome the agent consistent problem brought by GAT, a split loss is introduced to distinguish different agents. Results reveal that the proposed method achieves a significant improvement in the SMAC benchmark. Because of the group concept, our approach maintains excellent performance with an increase in the number of agents.

\end{abstract}

\section{Introduction}

The multi-agent system is widely applied in many real-life applications, including sensor networks, aircrafts formation flight, multi-robot cooperative control \cite{abs-1709-06011}, and networked autonomous vehicles. In these systems, cooperative multi-agents aim to complete a specific task via cooperating \cite{marl}. Most methods assume that agents can communicate in cooperation, which is impractical in some situations. Even though agents can communicate, the connection is unreliable due to interferences in environments. In this situation, the solution will be sub-optimal.

\begin{figure}[t]
\centering
\includegraphics[width=0.9\columnwidth]{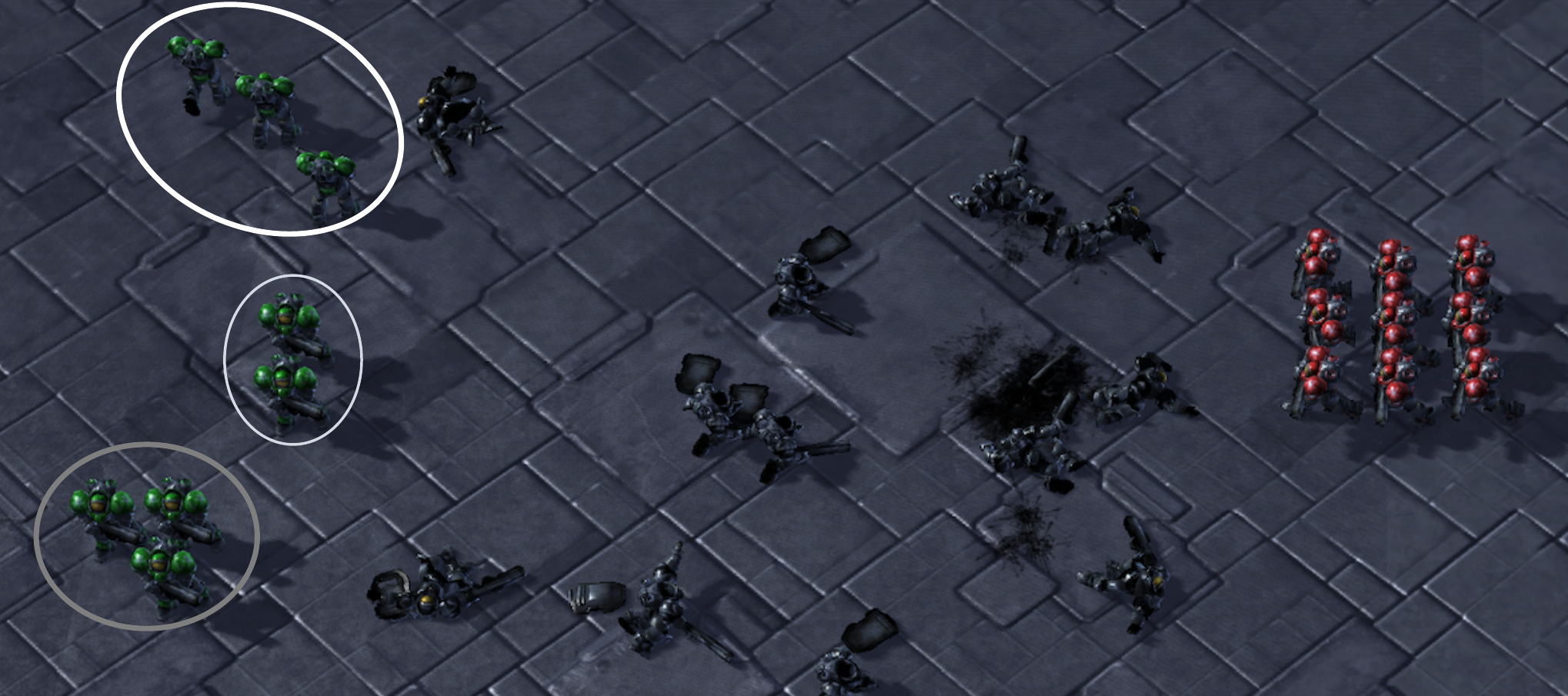} 
\caption{The process in SMAC via the Graph Clustering method. In this 8m\_vs\_9m map, agents are divided into three groups to complete the corresponding sub-tasks.}
\label{fig1}
\end{figure}

Many animal populations tend to work in groups to execute tasks, such as ants and geese. This hints that the agents in the same group may achieve better performance with less communication or even no communication. The reason is that agents in the same group tend to achieve similar tasks. Moreover, by dividing all agents into specific groups, the relationship between all agents is decomposed into the connection between groups, significantly reducing the complexity. Inspired by these, we propose a group concept for multi-agent reinforcement learning (MARL) to handle cooperative tasks without communication. Specifically, we assume that each agent has a belief to assist in decision making, and agents in the same group tend to generate similar beliefs.   

In the large-scale multi-agents scenario, it is challenging to model all agents in a limited time efficiently. A simple way is to use a bidirectional recurrent neural network to represent all agents as a group explicitly. Nevertheless, it ignores the second-order relationship information between agents and is limited by the sequence of agents. In this paper, we propose a straightforward approach to map agents. Inspired by the neighborhood cognitive consistency \cite{NCC}, we propose to use the k-Nearest Neighbor (kNN) method to map agent via agent-relative position. The idea is that agents in adjacent positions will have similar observation features, which leads to a similar decision. 

Our contribution has four parts. (1) We design a novel modular agent structure that can achieve the Centralised Training and Decentralised Execution (CTDE), compared to the communication-based method. (2) We propose to use the agent belief to represent the approximate group feature which is optimized via the GAT module and achieve approximate performance without communication. (3) We propose an explicit method to mapping agents to achieve similar beliefs in adjacent agents. (4) To overcome the consistent problem brought by GAT, we introduce a split loss to distinguish agents.

The proposed group-based method is evaluated on several unit micromanagement tasks based on StarCraft II \cite{SMAC}. The results indicate that our algorithm performs better than traditional ones in the literature. The performance of the method scales well with the number of agents. To observe the correlation between group features and agent's topology information, the t-SNE method \cite{T-SNE} is used. The experimental result reveals that the feature of agents in adjacent positions are also adjacent.

\section{Related Work}

Over the past years, deep multi-agent reinforcement learning has made a considerable breakthrough, widely used in games, traffic control, and other fields. This research concentrates on cooperative MARL with a value function-based method. In this setting, all teams receive a global team reward, so it is essential to divide the global team reward into individual rewards. \citet{VDN} decomposes the joint Q value into the individual Q value of each agent. \citet{QMIX} presents the constraint in which the joint Q value and the individual Q value have a monotonous setting based on the VDN algorithm. The suboptimal problem and decentralization in multi-agents are balanced by \citet{QTRAN} via an L2-penalty term.

Among current multi-agent algorithms, communication is a crucial point. The communication-based approaches assume that multiple agents can make essential information interactions to assist in deciding. \citet{RIALDIAL} applies a deep feedforward neural network to generate communication vectors for agents' communication. \citet{BICNET} utilizes a bidirectional recurrent neural network for communication between multiple agents. \citet{LIIC} proposes a selectable point-to-point communication method adopted to determine whether agents communicate with each other by constructing a belief vector. Unlike these methods, \citet{GCRL} employs graph convolution network and multi-head dot-product attention to aggregate agent features. Because agents generate their actions based on other agents' beliefs or relative features, these methods always need centralized execution that follows the CTCE framework. However, when the actual scenario restricts some agents from being difficult to communicate, the agent may obtain inaccurate beliefs or wrong beliefs, leading to sub-optimal actions. 

Current researches on agent mapping always focus on the attention mechanism. \cite{MAGA} utilizes two-stage attention to build an adjacent matrix. Hard attention is devoted to delete related weak edges, and soft attention generates the weight coefficients of the retained graph structure. EPC-MADDPG \cite{long2020evolutionary} uses the scalar dot product attention to fuse the variable entity features to achieve a global mapping. However, simply using the attention mechanism to capture the association between agents for mapping will cause the algorithm to fall into a locally optimal solution. If the initial attention mechanism considers that two far apart agents are similar and tend to similar tasks, all agents will behave the same. All agents will consistent and lose the meaning of grouping.

\begin{figure*}[htbp]
\centering

\subfigure[The schematics of our approach]{
\begin{minipage}[t]{0.55\textwidth}
\centering
\includegraphics[width=4in]{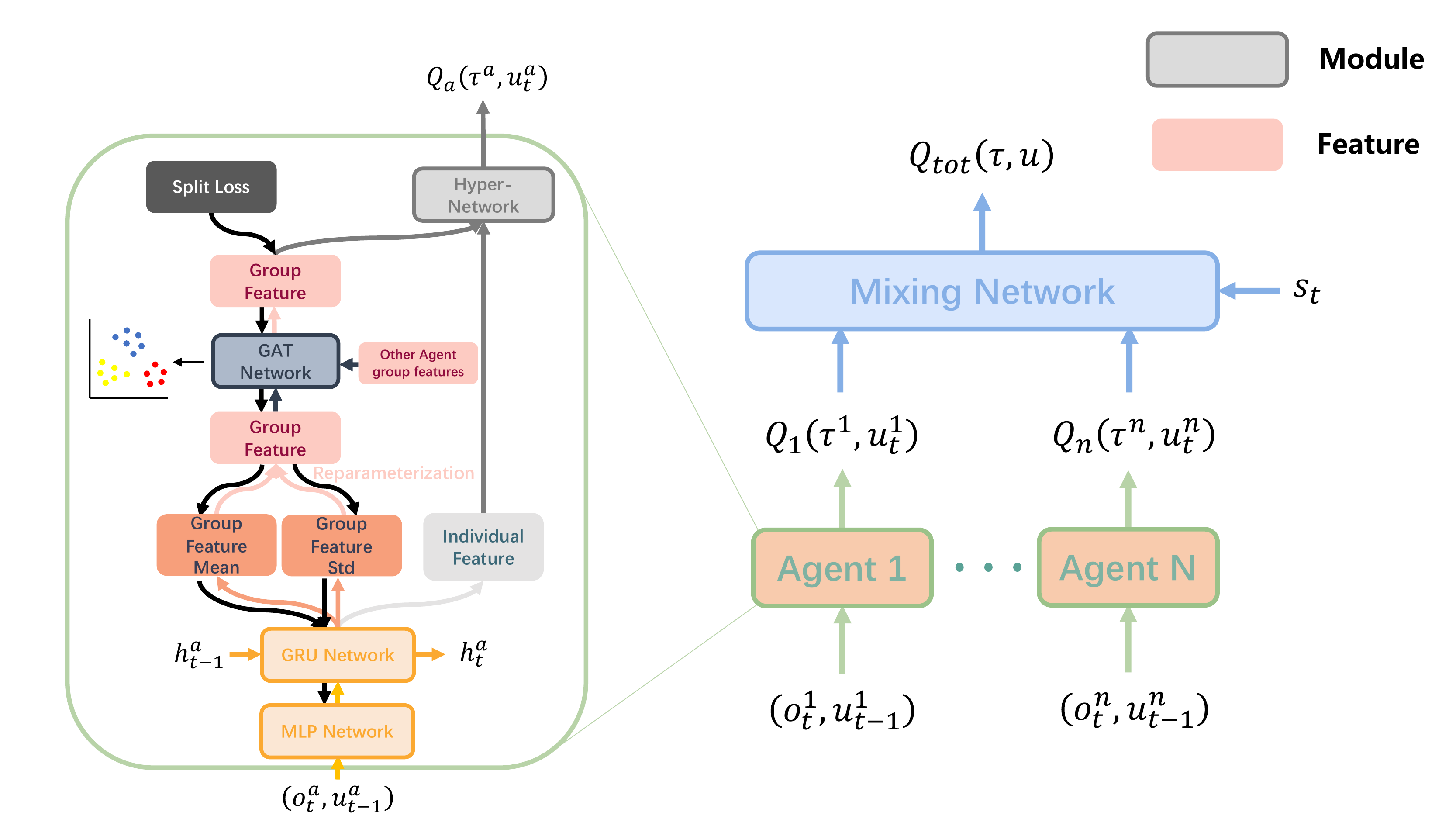}
\end{minipage}}
\subfigure[The schematics of our approach base on the distributed framework]{
\begin{minipage}[t]{0.4\textwidth}
\centering
\includegraphics[width=2.8in]{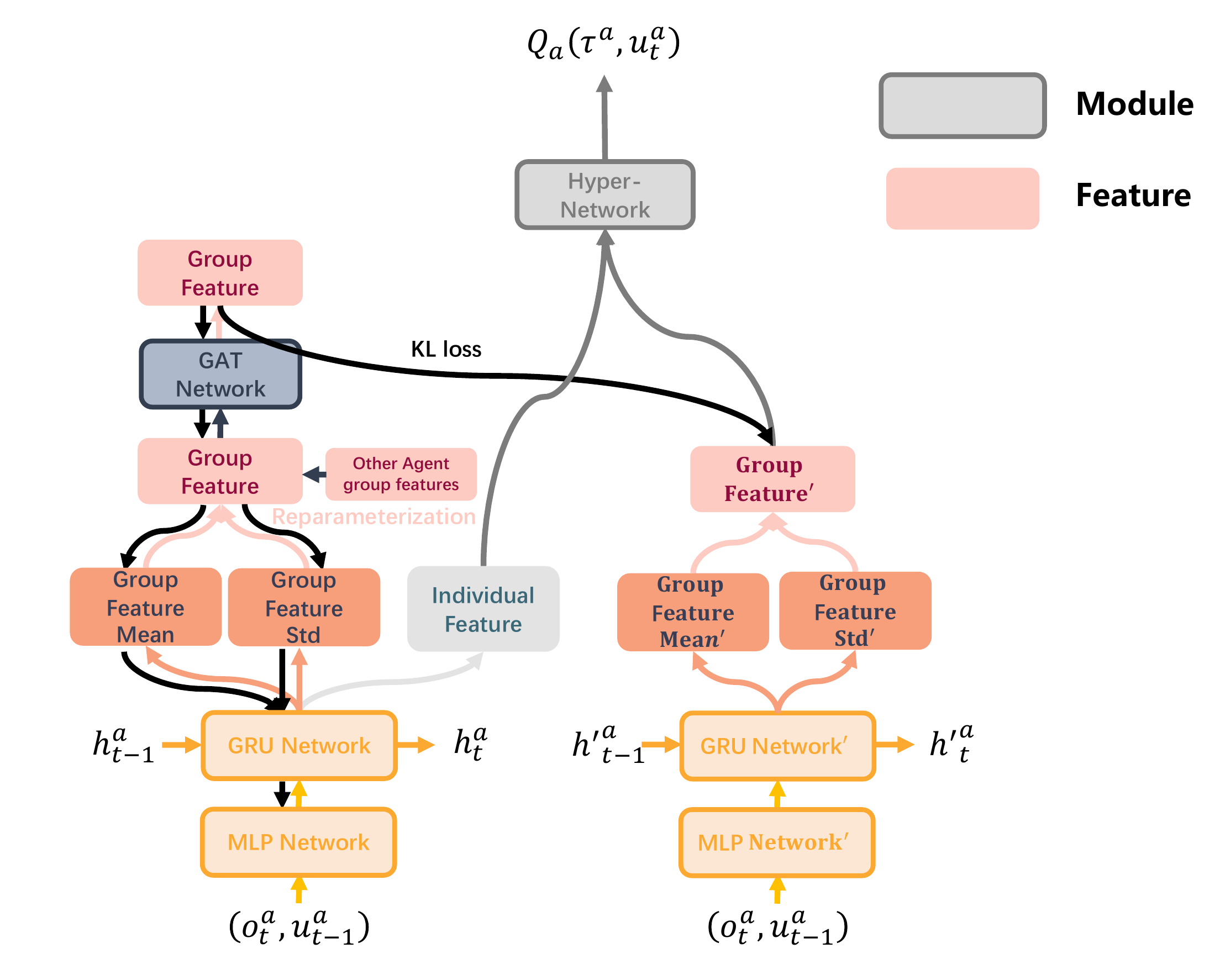}
\end{minipage}}  
 
\centering
\caption{(a) The Encoder generates individual features and Gaussian distribution's group features. Then group features are sampled via reparameterization trick and are fed into the GAT network to make feature aggregation. The hypernetwork merges the group features and the individual features into the individual Q value to generate the total Q value via QMIX. The black line indicates the route of gradient propagation of the split loss. (b) Our algorithm can be decomposed into modules. Specifically, we use a new group feature network instead of the original group network to achieve the purpose of distributed exection. The new group feature network is independent of other agents. This new network is trained through minimizing the KL divergence between the group feature network's output and the original group network's output.}
\label{fig2}
\end{figure*}


\section{Background}
In the present research, the problem is regarded as a fully cooperative multi-agent task viewed as a Dec-POMDP \cite{ctde1} comprising of a tuple $\left< I, S, U, Z, P, R, O, n,\gamma\right>$, where $s \in S$ depicts the global state of the environment. At any time, each agent $i \in I \equiv \{1,...,n \}$ interacts with the environment by generating corresponding actions $u_i \in{U}$ through the observation vector $z_i \in Z$ according to the observation function $O(s, i)$. Agents learn to maximize the reward $R$ for environmental feedback. This process is based on a state transition function $P\left(s^{\prime} \mid s, a\right)$. Moreover, $n$ signifies the number of agents, and $\gamma$ represents a discount factor.

Interaction-based MARL algorithms are generally implemented via the centralized training with centralized execution (CTCE) framework. Agent's strategy $\pi_i(u_i|\tau_i)$ is generated based on the observation sequence $\tau$, the global state $s$, and the interactive features of other agents. Via introducing the modular mechanism, our framework is designed as the distributed and applying a regularization tool to generate representative group features without interaction.

\subsection{Graph Attention Network}
Graph Attention Network (GAT) can correlate similar features between agents using masked self-attentional layers. To better correlate similar features between agents, GAT uses masked self-attentional layers. By introducing the attention mechanism, the neural network can focus on the most relevant parts of the input, which helps learn the correlation features adaptively between nodes \cite{gat}. Besides, GAT can capture relative features between disconnect nodes by stacking GAT layers. GAT has two attention mechanism types, the Global Graph Attention and Masked Graph Attention. The former builds the attention operation between each one node and all the other, while the latter only performs the same operation on neighboring nodes.

Distant agents may bring a negative effect on the current agent. Also, it is very computationally expensive to calculate the relationship between all agents, which causes the Global Graph Attention-based method unreasonable in large-scale scenarios. While Masked Graph Attention-based method only captures the relationship between the adjacent agents, which leads neighbor agents to hold similar actions. Besides, agents can capture the unconnected agent features via stacking the Masked Graph Attention layers and broaden the receptive field of agents.


\section{Method}
In this section, we introduce the group concept into the multi-agent algorithm and propose the Belief in Graph Clustering (BGC) methods. The overall framework of the algorithm is shown in Figure \ref{fig2}. To obtain a distributed framework, a modular approach is applied to build an algorithm network and generate groups and individual features of agents separately. Due to the idea of modular construction, we utilize a new group feature network to represent agent group features via minimizing the difference of original group features without agent intersection. We use the group features to approximate the effect of communication to implement the distributed execution. 

In the overall pipeline, each agent generates an individual Q value via its local observation and then passes it into a mixed network (such as QMIX) for generating the global Q value. Specifically, each agent encodes its local observation sequence via a GRU network to produce agent group features and individual features. The group features are obtained via exercising the multivariate Gaussian distribution condition on the observation sequence. The reparameterization trick is practiced for sampling to ensure the continuity of the gradient. Then, to cluster group features and obtain relationship features from adjacency agents, a GAT network is adopted. Group and individual features are feeds into a hyper network \cite{hypernetwork} to generate individual Q Value. Besides, to prevent all group features from converging together, we introduce a split loss to distinguish non-adjacent agents. 

\subsection{Adjacent Matrix via kNN}
In this section, we introduce a k-Nearest Neighbor (kNN) method to represent agent relationships on topology and construct the adjacency matrix for the masked attention-based GAT network. Due to the centralized training setting, it is easy to get all agents' position information from agent observation. Agent adjacency graph is constructed by defining the agent with the k nearest agents as the same type. Then we take the Laplace transform and regularization of adjacent graphs to produce the agent adjacency matrix.

In our setting, the hyperparameter k in kNN is two. In this setting, all agents will be divided into several groups, as shown in Figure \ref{fig5}(d). The non-information exchange between the group level leading different groups to achieve different sub-tasks to realize group diversity. Besides, when all agents are adjacent, it leads the group-level information transfer, which helps accomplish the finished task via group coordination.

\subsection{Belief in Graph Clustering}
In this part, we introduce the details on how to generate agent belief via graph clustering. First, we embed agent observation via an MLP network. To capture the relevant features on time series, we introduce the GRU module. To add the uncertainty to agent group features and enhance the exploration ability, we use a Gaussian Sampling module, which takes the condition on the embedding features of the GRU module. This module uses  MLP networks to generate the mean and variance (take the exponential function on the logarithmic variance term) to construct the independent Gaussian distribution. Besides, we introduce the reparameterization trick to sample the actual group features for the gradient's continuity.

\begin{gather}
\left(\mu_{g_{i}}, \sigma_{g_{i}},s_i \right)=f\left(\tau_{i} ; \theta_{i}\right) \tag{1a}\\
g_{i}=\mu_{g_{i}}+\sigma_{g_{i}} \odot \varepsilon_{i} \quad \varepsilon_{i} \sim N(0,1) \tag{1b}
\end{gather}
where $\theta_i$ represents parameters of network, $\mu_{g_{i}}$ and $\sigma_{g_{i}}$ signify the mean and standard deviation of group features, $s_i $ represents individual features, $g_{i}$ signifies the actual group feature, and $\varepsilon_{i}$ is noise.

In centralized training, agents can use some adjacent agent features to assist in deciding. After getting the adjacency matrix, we utilize the GAT module to cluster the relevant features into group features. Under this setup, the agent interacts with agents in the adjacent position via the attention mechanism and generates the correlation weight between agents. This weight is used to calculate the agent relationship features for the current agent, and the agent fuses these features to produce final group features.

\begin{gather}
	g_{i}^{\prime}=\sigma\left(\sum_{j \in \mathcal{N}_{i}} \alpha_{i j} g_{j}\right) \tag{2}
\end{gather}

where $\sigma(\cdot)$ is the ReLu activation function, and $\alpha$ is the clustering coefficient calculated via the function. 
	
\begin{gather}
	e_{i j}=  a\left(\left[W g_{i} \| W g_{j}\right]\right), j \in \mathcal{N}_{i} \tag{3a} \\ 
    \alpha_{i j}=  \frac{\exp \left(\text {LeakyReLU}\left(e_{i j}\right)\right)}{\sum_{k \in \mathcal{N}_{i}} \exp \left(\text {LeakyReLU}\left(e_{i k}\right)\right)} \tag{3b}
\end{gather}

Finally, we use a hyper network module to fuse group features and individual features to generation agent action. In this module, the group features are utilized to generate the MLP network weight, multiplying individual features to get the final action.

\subsection{Split Loss}
Although mask attention-based GAT network can prevent all agents from forming a single group, current agents' features will still extend to all other agents due to the time perspective. Therefore, we propose a split loss to alleviate this issue. We use Kullback-Leibler (KL) divergence to measure differences between agents' group features and keep the non-adjacent agents at a fixed distance.

\begin{align}
	 \max_{\pi \in \Pi} \quad &  \mathbb{E}_{\tau \sim \rho_{\pi}}\left[\sum_{t=0}^{T} r\left(\mathbf{s}_{t}, \mathbf{a}_{t}\right)\right] \tag{4a} \\ 
    s.t. \quad & \mathbf{KL}(g_i||g_j) \geq \delta, \forall \ edge(i, j)=\phi \tag{4b}
\end{align}

where $g_i$ represents agent group features and $edge(i, j)$ determines whether there is a connection between agent $i$ and agent $j$. 
	 
Therefore, a split loss is introduced to separate the agent: 
\begin{gather}
	     L_{split}=-\sum_i^N \sum_j^N min(\mathbf{KL}(g_i||g_j)-\delta, 0), \notag \\
    \forall \ edge(i, j)=\phi ,i \neq j \tag{5}
\end{gather}

The hyperparameter $\delta$ is employed to keep agents at the $\delta$ distance, changing with the scenario.

\subsection{Decentralization Execution}

To realize the complete decentralized execution without agent interaction, we apply a new sub-network to learn the new group features and use the KL divergence to minimize the distance between the new group features and the original group features learned by the GAT network. Then, the gradient will flow through the new sub-network and train the new sub-network. This frame can be view as a teacher-student structure.


\begin{figure*}[htb]
\centering
\begin{minipage}[t]{0.33\textwidth}
\centering
\includegraphics[width=1.1\columnwidth]{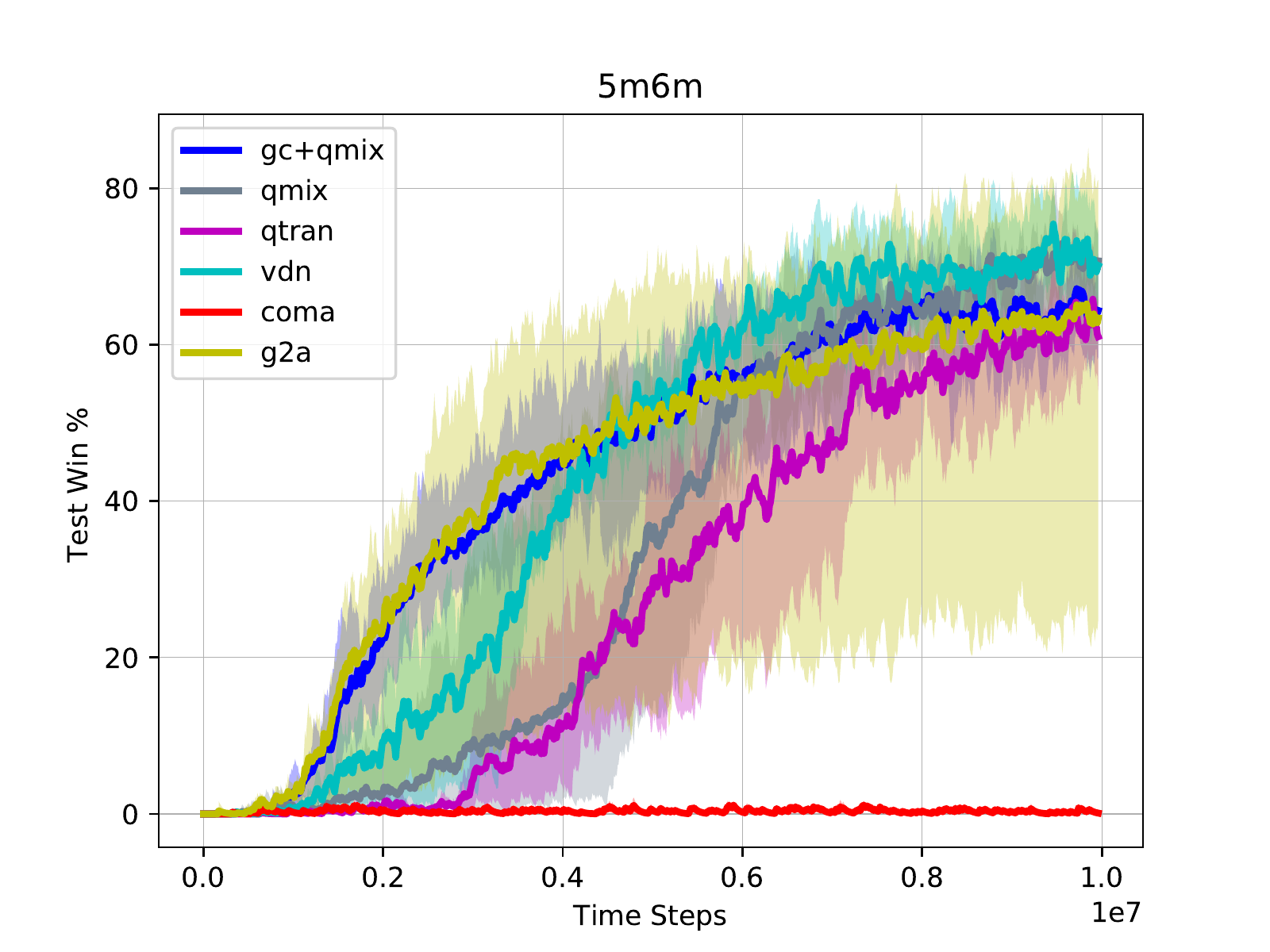}
\end{minipage}
\begin{minipage}[t]{0.33\textwidth}
\centering
\includegraphics[width=1.1\columnwidth]{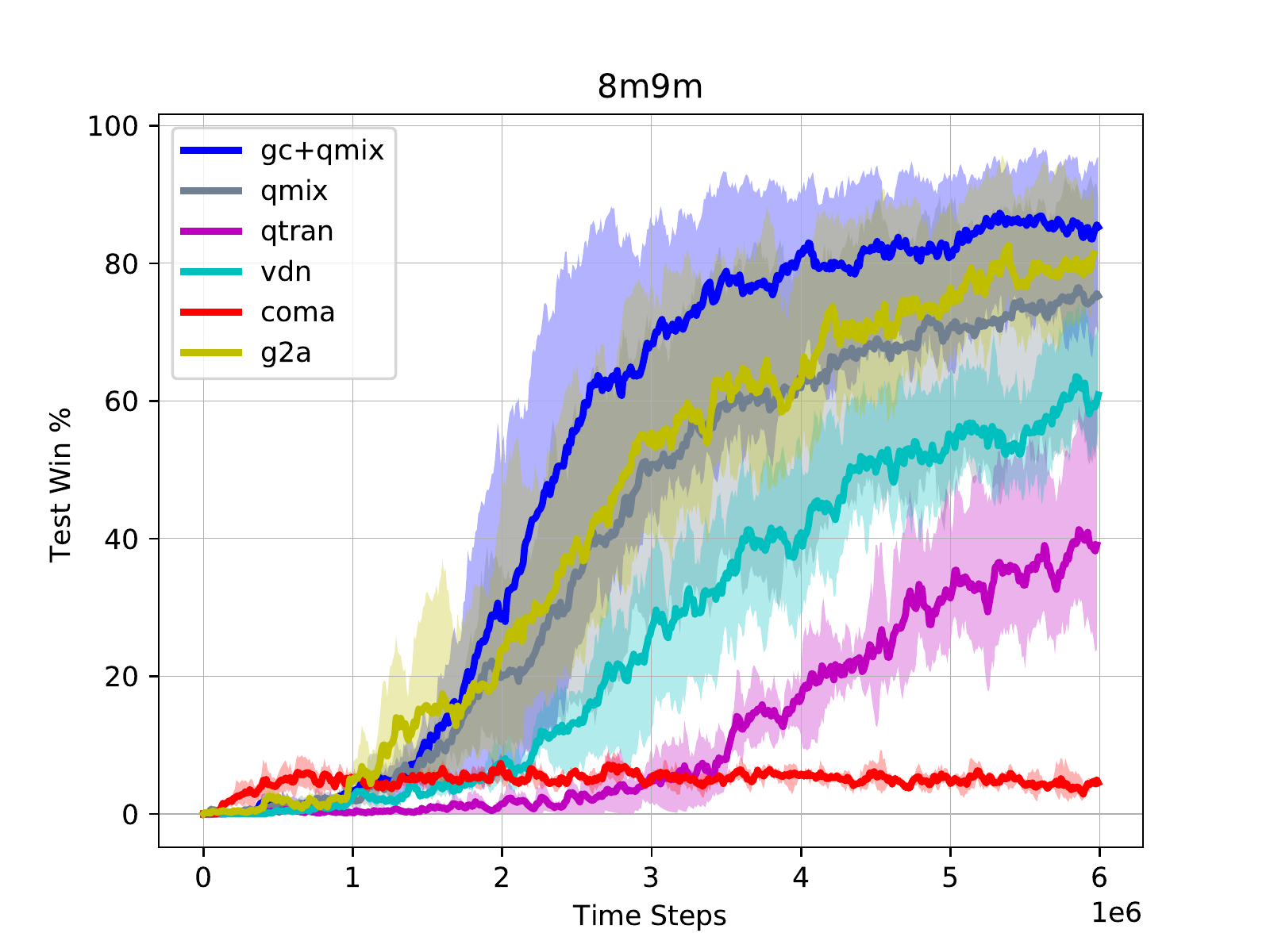}
\end{minipage}
\begin{minipage}[t]{0.33\textwidth}
\centering
\includegraphics[width=1.1\columnwidth]{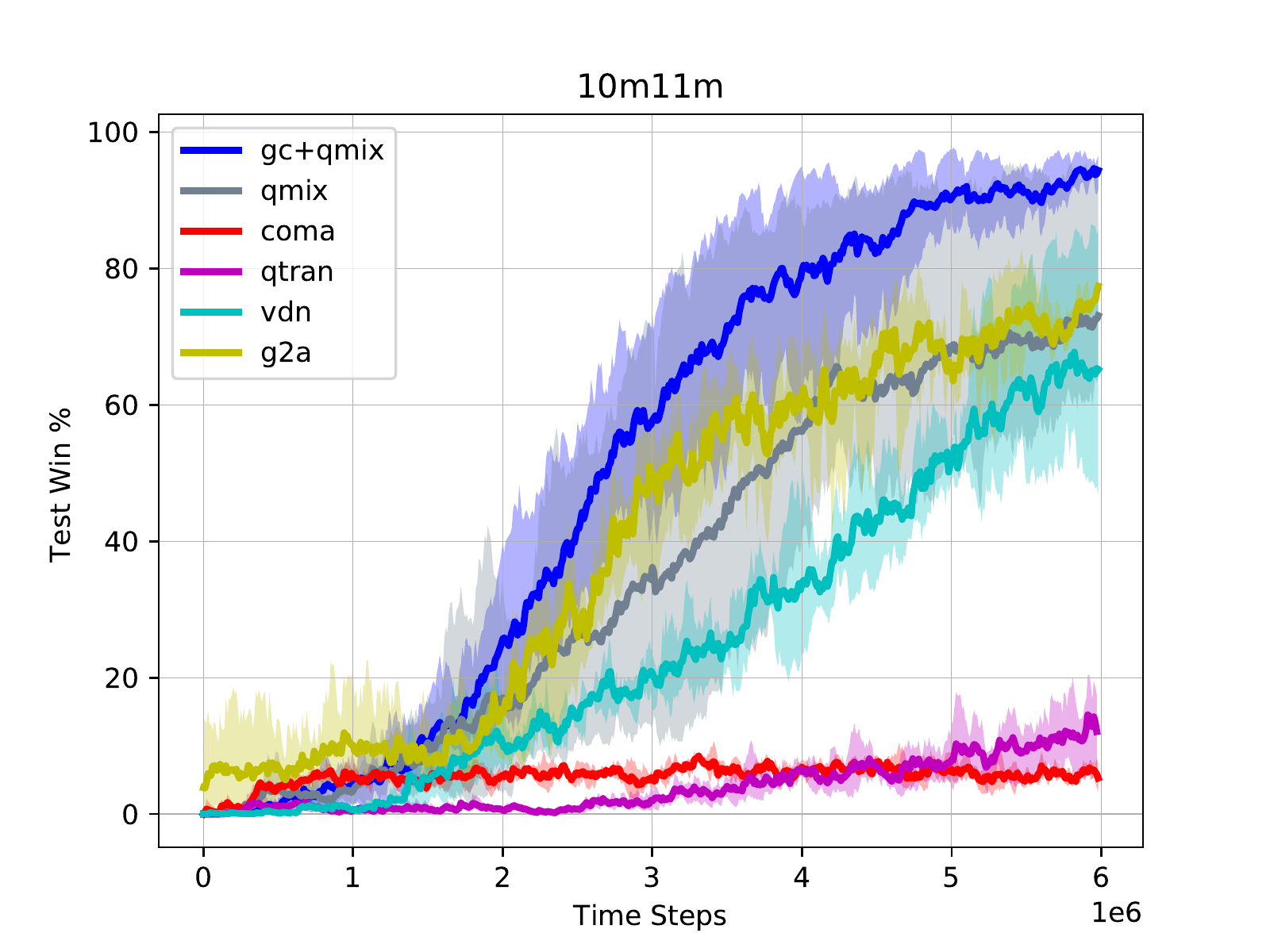}
\end{minipage}
\centering
\begin{minipage}[t]{0.33\textwidth}
\centering
\includegraphics[width=1.1\columnwidth]{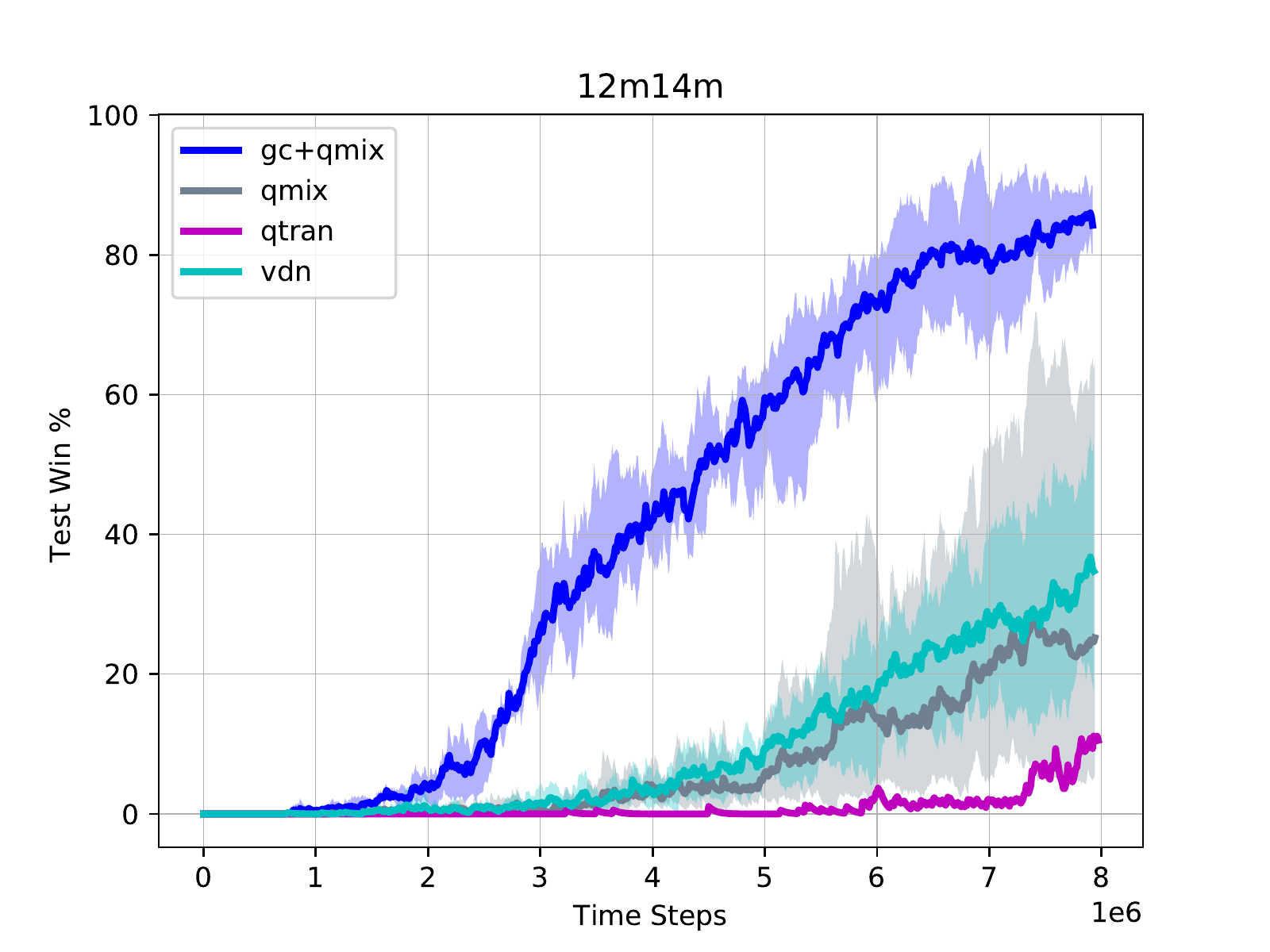}
\end{minipage}
\begin{minipage}[t]{0.33\textwidth}
\centering
\includegraphics[width=1.1\columnwidth]{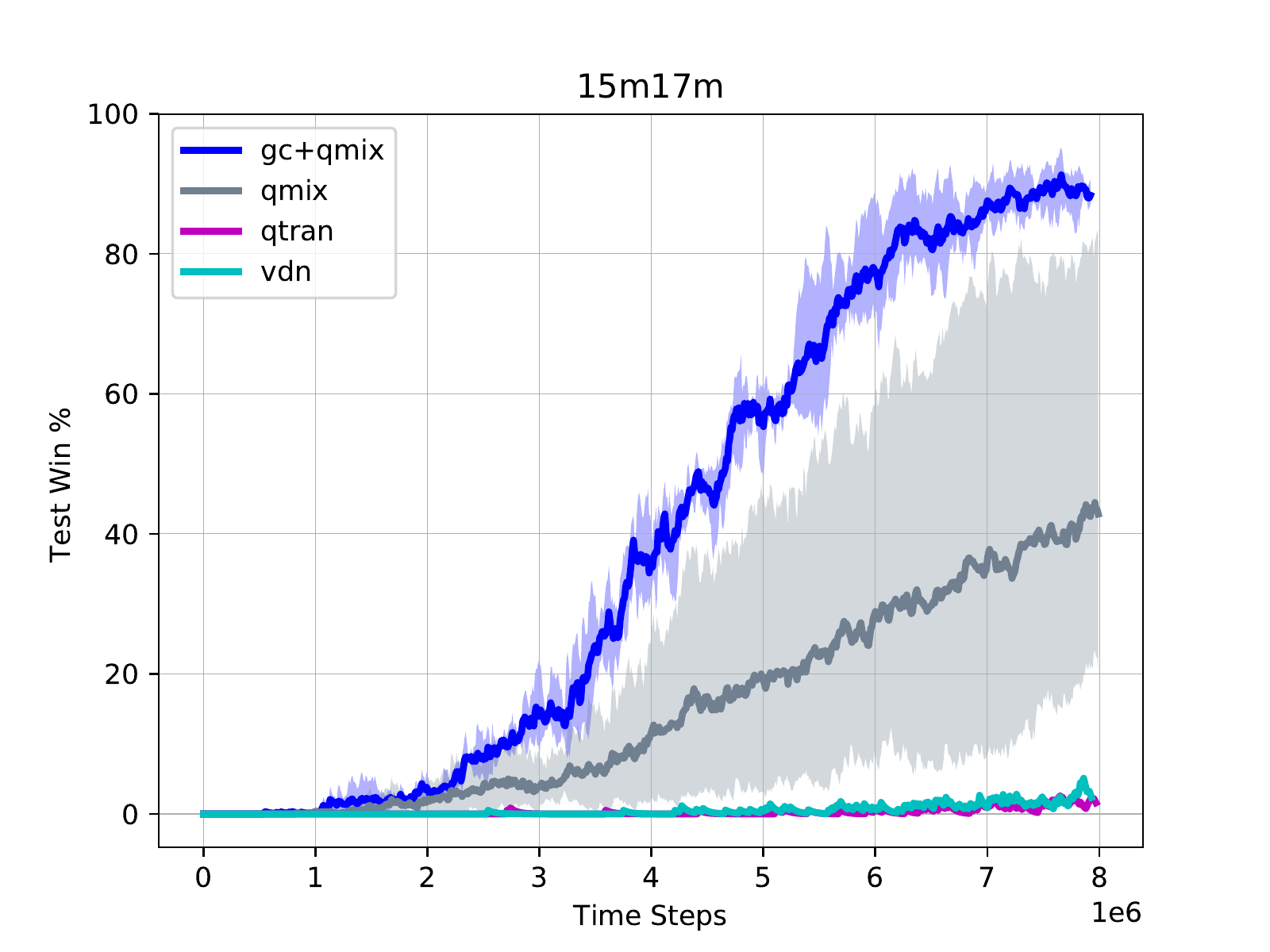}
\end{minipage}
\begin{minipage}[t]{0.33\textwidth}
\centering
\includegraphics[width=1.1\columnwidth]{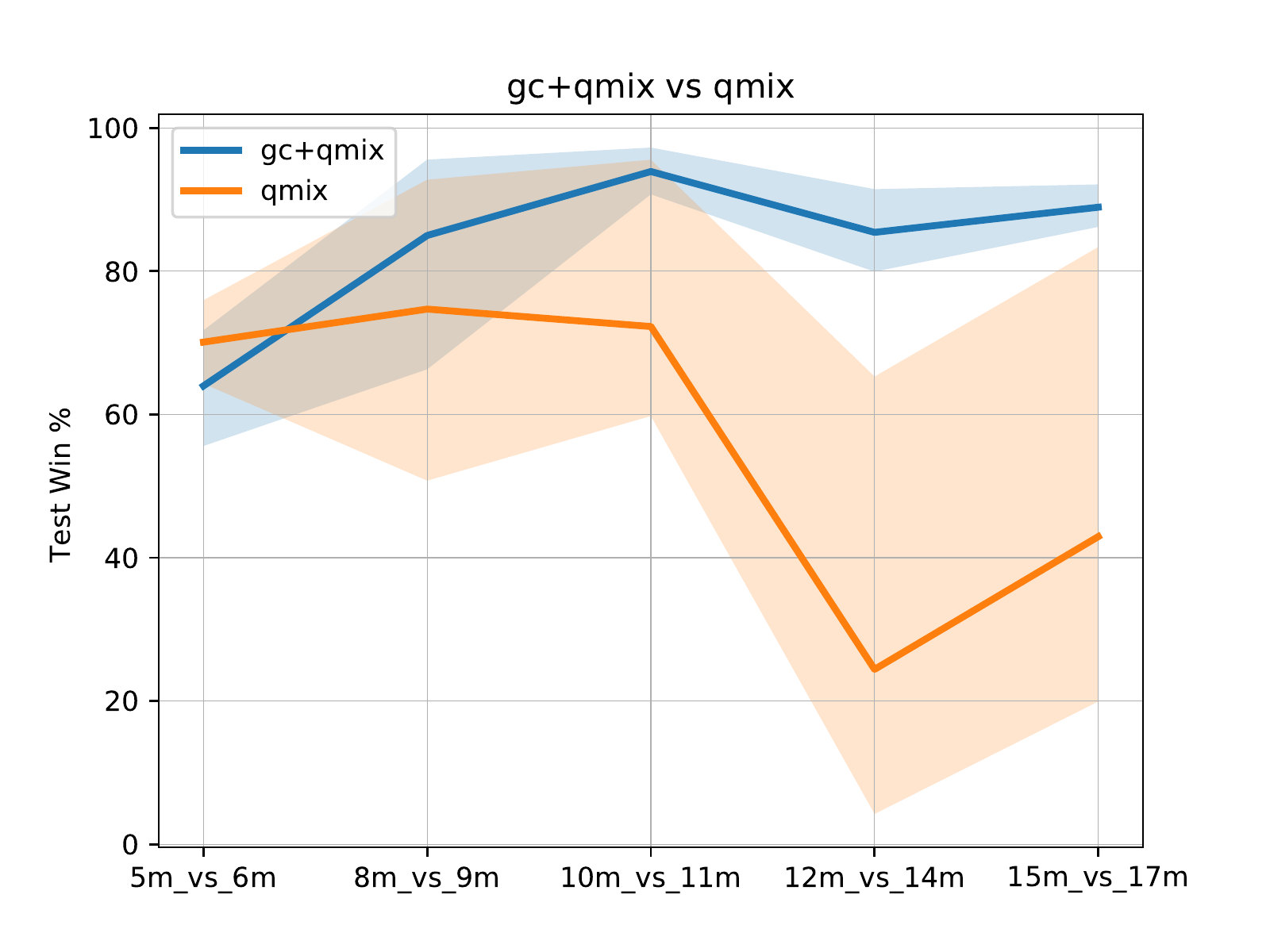}
\end{minipage}
\caption{Result of starcraft II. The comparison between the group-based method and other algorithms with the number of agents increasing from 5 to 15. Empirical results show that as the number of agents increases, our algorithm show more superiority comparing with the baseline algorithm. Due to the limitation of graphics card memory(RTX2080Ti), we did not test the actual effect of G2ANet in the 12m\_vs\_14m map and 15m\_vs\_17m map.}
\label{fig3}
\end{figure*}

\section{Experiment}

\subsection{Starcraft II}
In the experiment part, we test our algorithm in the SMAC benchmark. The difference between SMAC and the traditional Starcraft II environment is that it focuses on unit micromanagement. It leverages the natural multi-agent microstructure by proposing a modified version of the problem designed specifically for decentralized control. To test the algorithm robustness, we run our algorithm in 6 to 10 million steps in 4 random seeds. Besides, to estimate the performance of the method, we experiment in five scenarios, three standard maps of 5m\_vs\_6m (very hard), 8m\_vs\_9m (very hard), and 10m\_vs\_11m (very hard), and two custom maps of 12m\_vs\_14m (very hard) and 15m\_vs\_17m (very hard).

Figure \ref{fig3} illustrates the performance difference between the proposed algorithm and other benchmark algorithms in five scenarios. We first compare our method with the origin QMIX algorithm. The result shows that BGC-QMIX is outperforming the origin algorithm on the final test win rate and the performance variance, as shown in Figure \ref{fig3} (BGC+QMIX vs. QMIX).

Except for the scenario with a small number of $5m\_vs\_6m$(very hard), our algorithm demonstrates a significant improvement compared to the baseline methods. In the $5m\_vs\_6m$ scenario, the proposed algorithm has excellent advantages over other ones in terms of convergence speed. In this scenario set, the small number of agents causes agents' group features to be consistent, leading to less than ideal performance. To check the influence of the number of agents, we further perform our method in other scenarios. \ref{fig3} (bottom right) shows that it can maintain an excellent performance with the number of agents increases while QMIX does not. The results show that the mask attention-based GAT method and the split loss performance.

Furthermore, our algorithm is compared with G2ANet, which is also based on the attention mechanism. G2ANet practices the hard attention to determine the communication target and soft attention to complete the communication. G2ANet overlooks agent with topology information's association using a bi-LSTM network to construct the hard attention operation. Further, the G2ANet is tested based on the QMIX framework, and the empirical result shows that our graph clustering-based algorithm performs better. (As the tests were under the hardware condition of RTX 2080Ti, $12m\_vs\_14m$ and $15m\_vs\_17m$ scenarios can cause cuda memory exhausted. Thus, only three standard scenes are tested.)


\begin{figure*}[htbp]
\centering
\begin{minipage}[t]{0.33\textwidth}
\centering
\includegraphics[width=1.1\columnwidth]{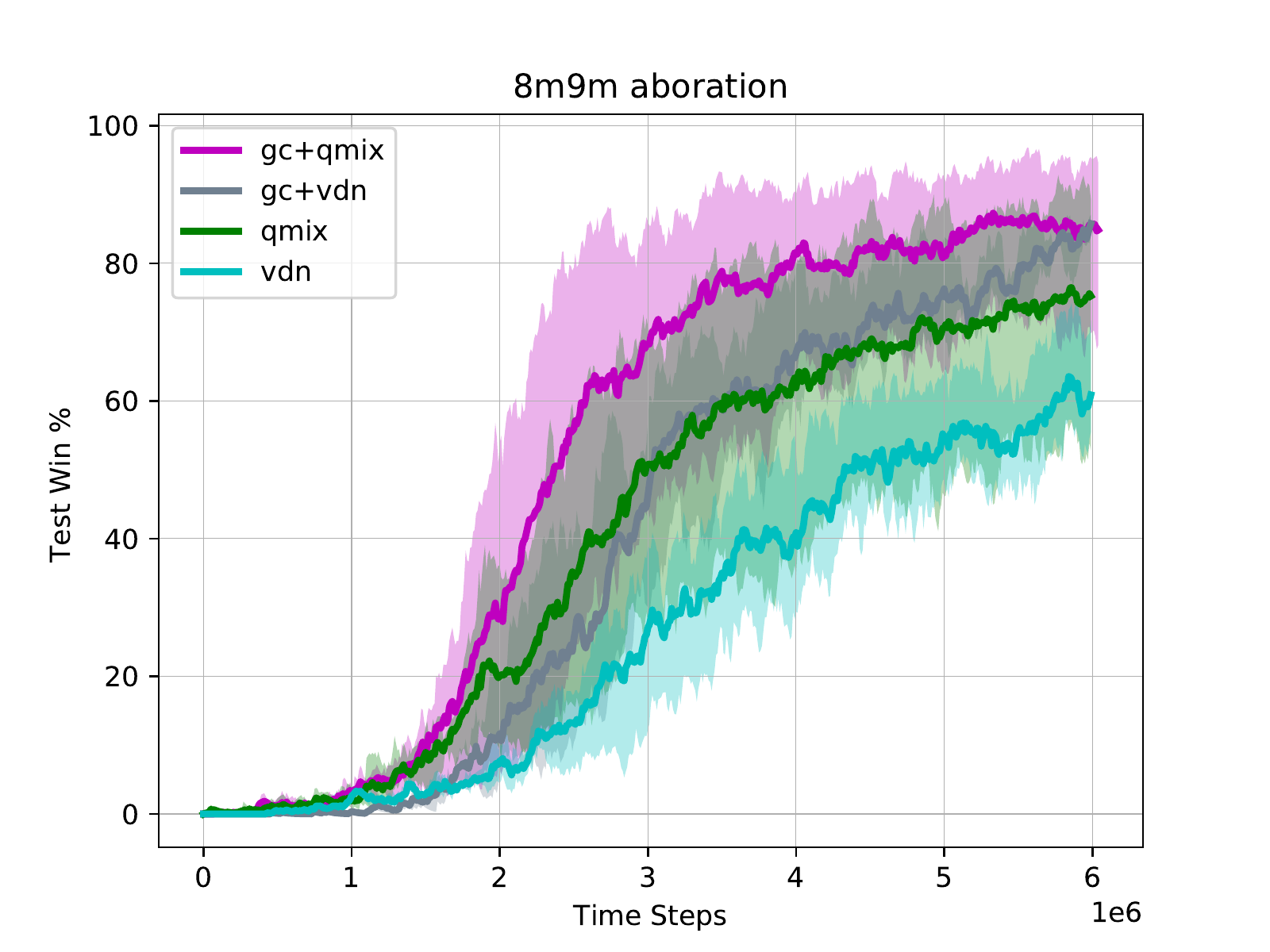}
\end{minipage}
\begin{minipage}[t]{0.33\textwidth}
\centering
\includegraphics[width=1.1\columnwidth]{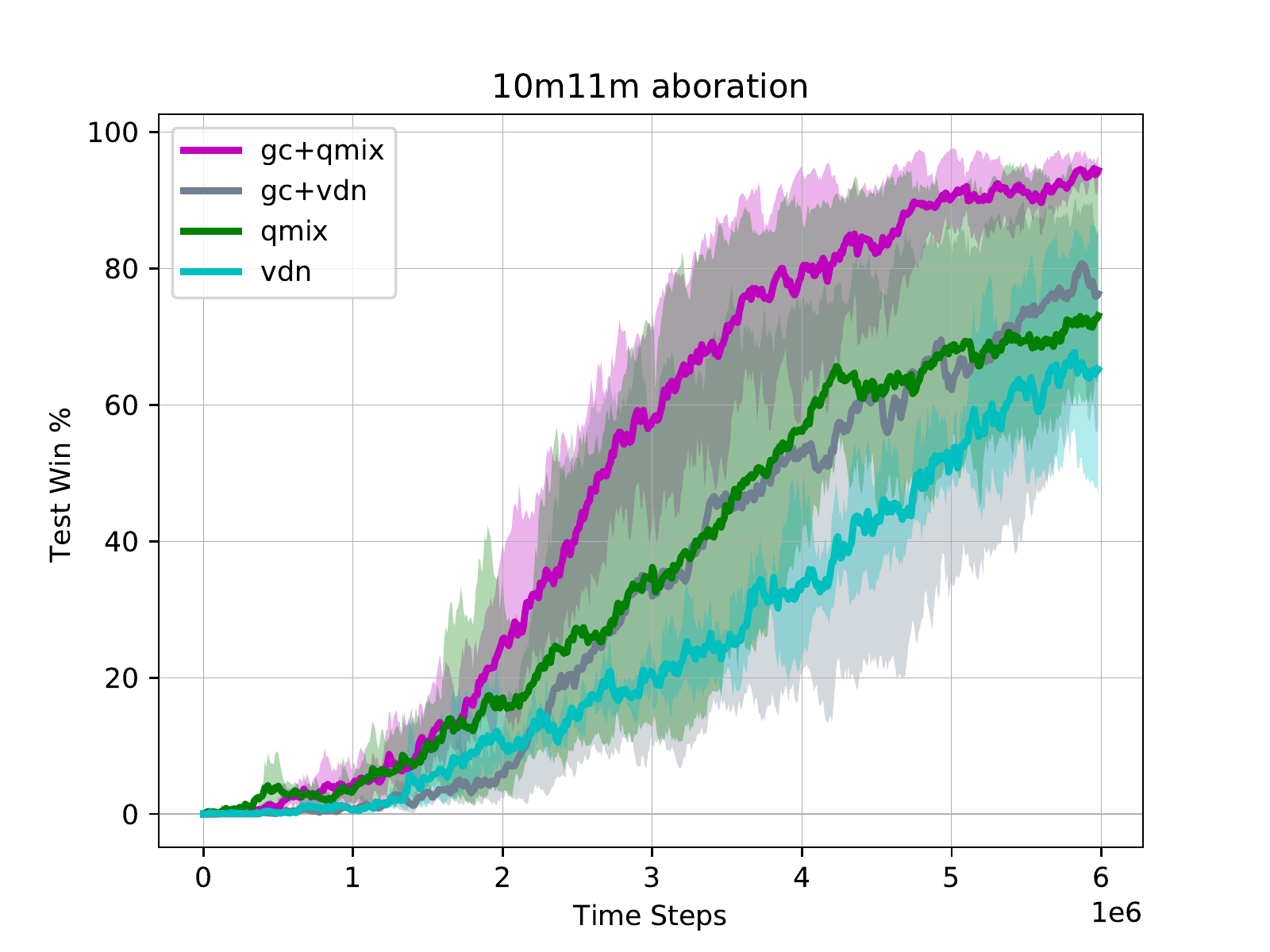}
\end{minipage}
\begin{minipage}[t]{0.33\textwidth}
\centering
\includegraphics[width=1.1\columnwidth]{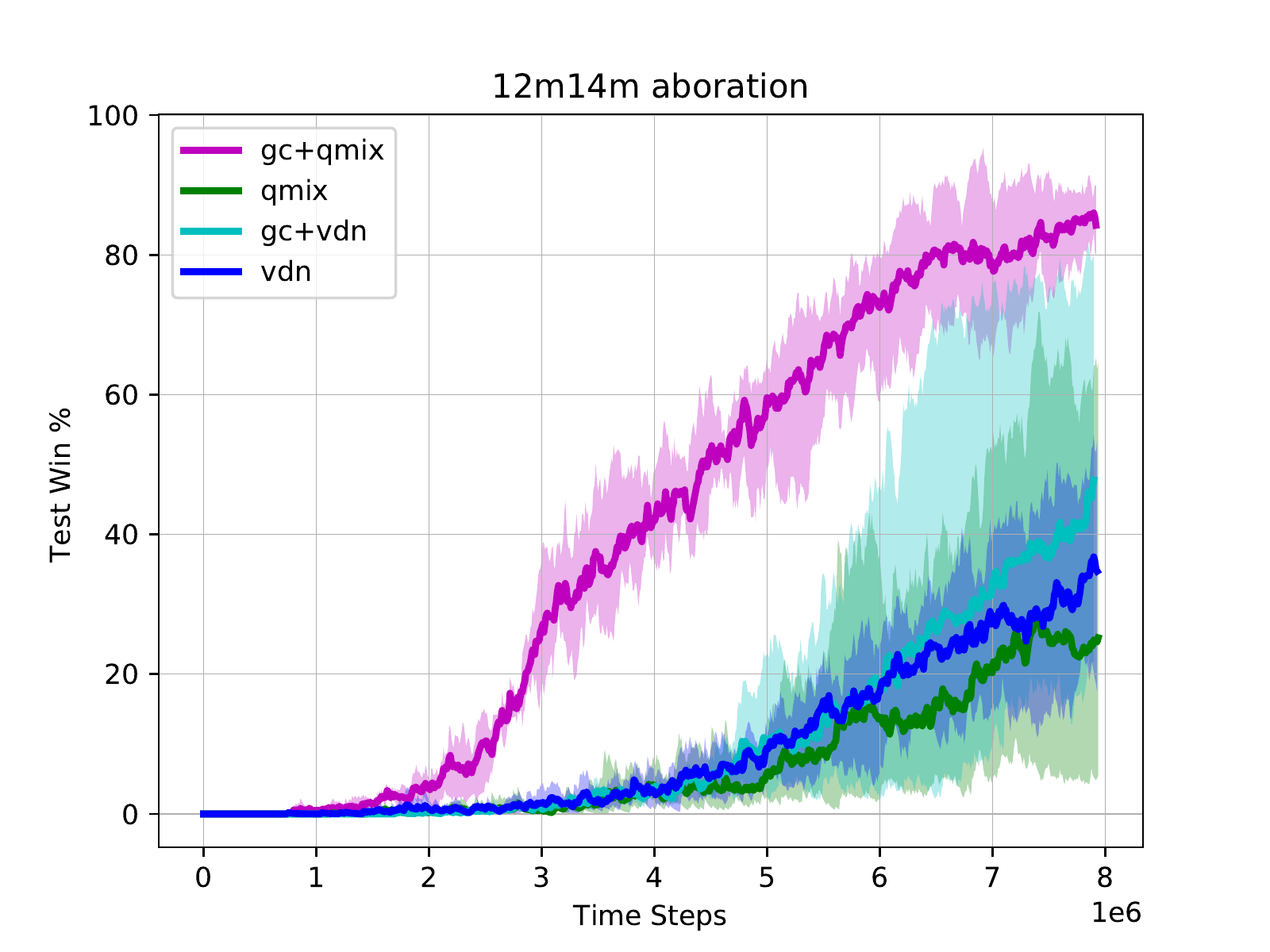}
\end{minipage}
\caption{Ablation test. we test the our graph clustering-based method base on the VDN framework. The result shows its superiority.}
\label{fig4}
\end{figure*}	

\begin{figure}[htbp]
\centering

\subfigure[8m\_vs\_9m(a)]{
\begin{minipage}[t]{0.2\textwidth}
\centering
\includegraphics[width=1.3in]{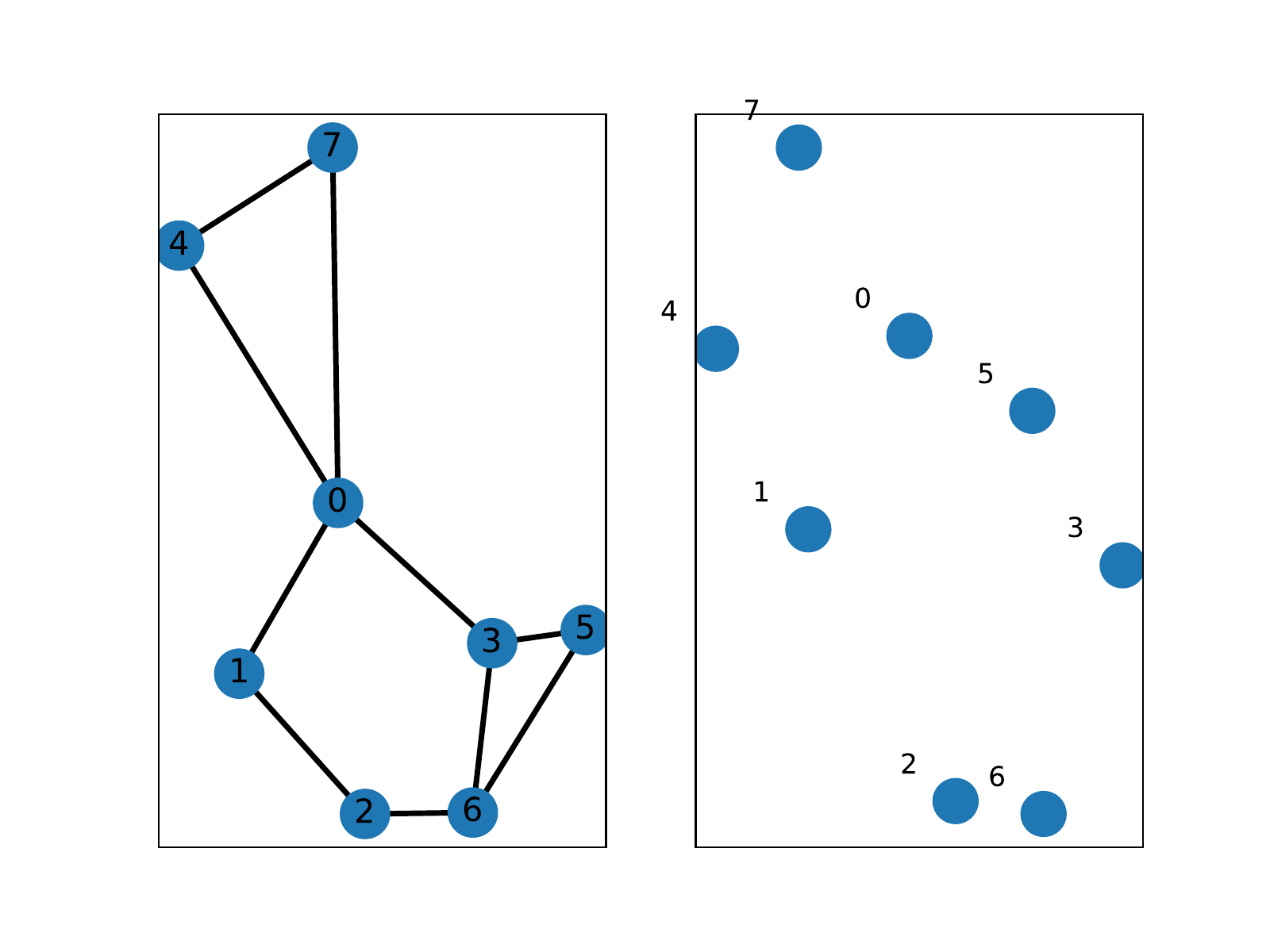}
\end{minipage}}
\subfigure[8m\_vs\_9m(b)]{
\begin{minipage}[t]{0.2\textwidth}
\centering
\includegraphics[width=1.3in]{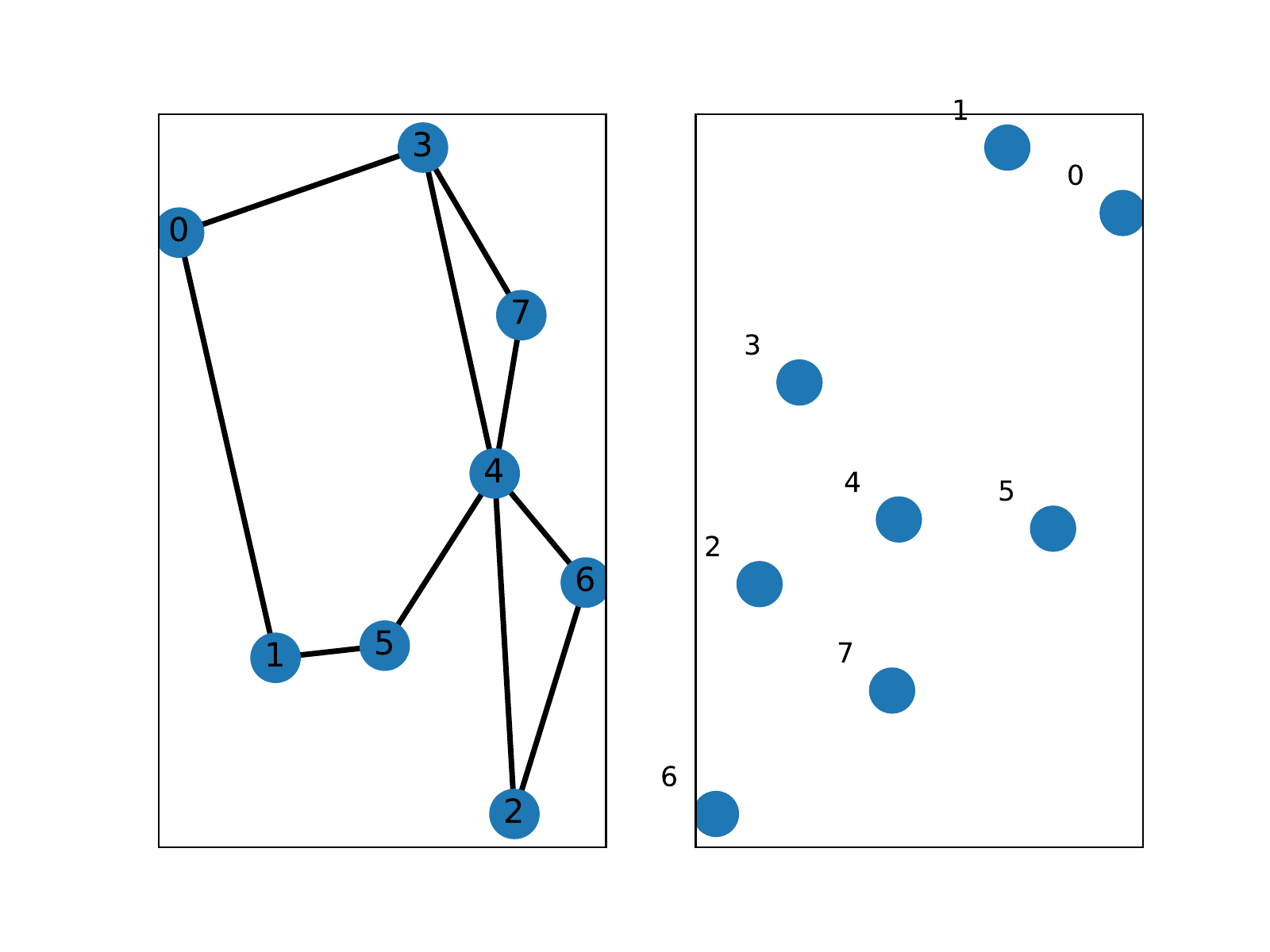}
\end{minipage}}  

\subfigure[10m\_vs\_11m(a)]{
\begin{minipage}[t]{0.2\textwidth}
\centering
\includegraphics[width=1.3in]{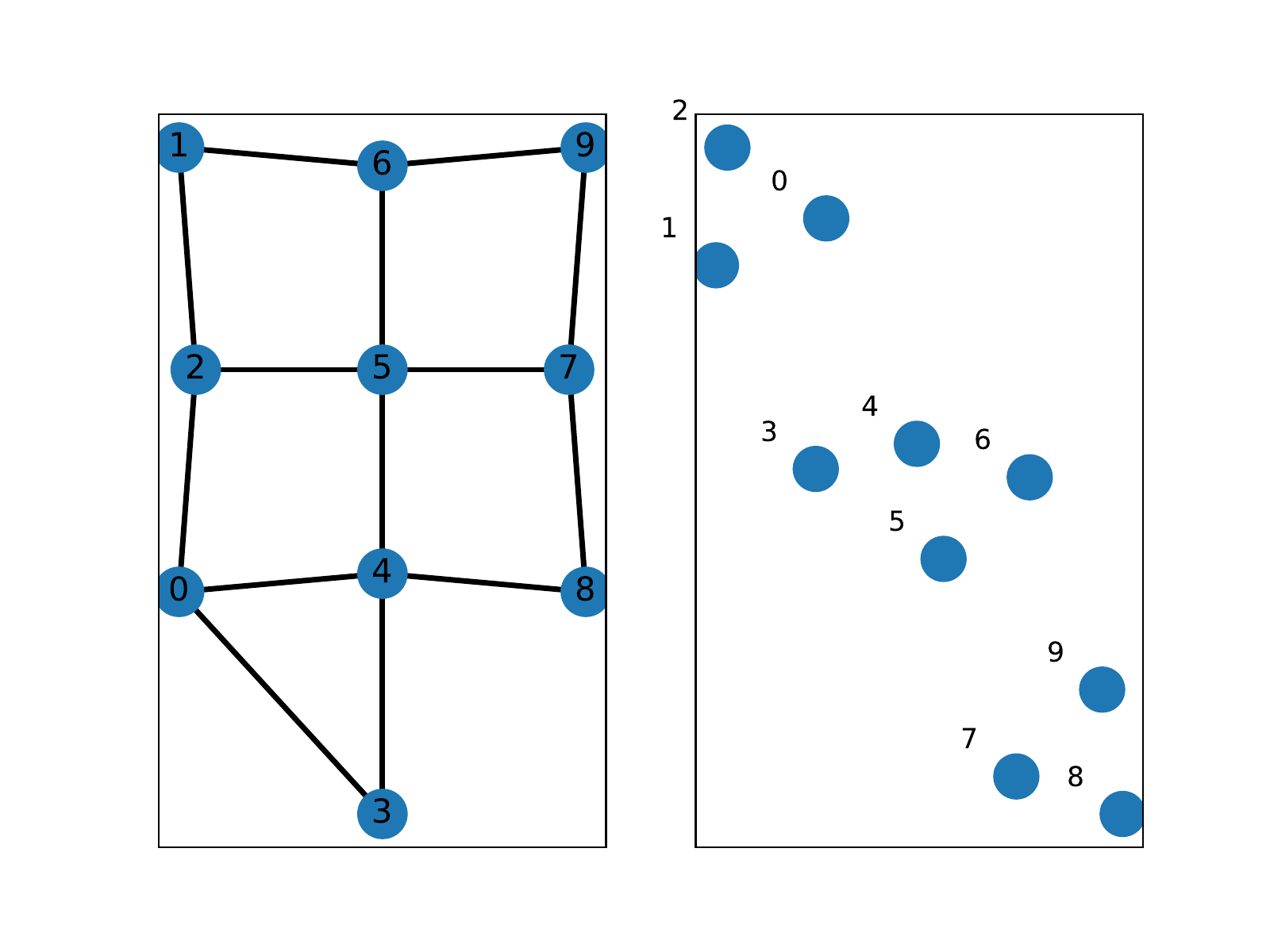}
\end{minipage}}
\subfigure[10m\_vs\_11m(b)]{
\begin{minipage}[t]{0.2\textwidth}
\centering
\includegraphics[width=1.3in]{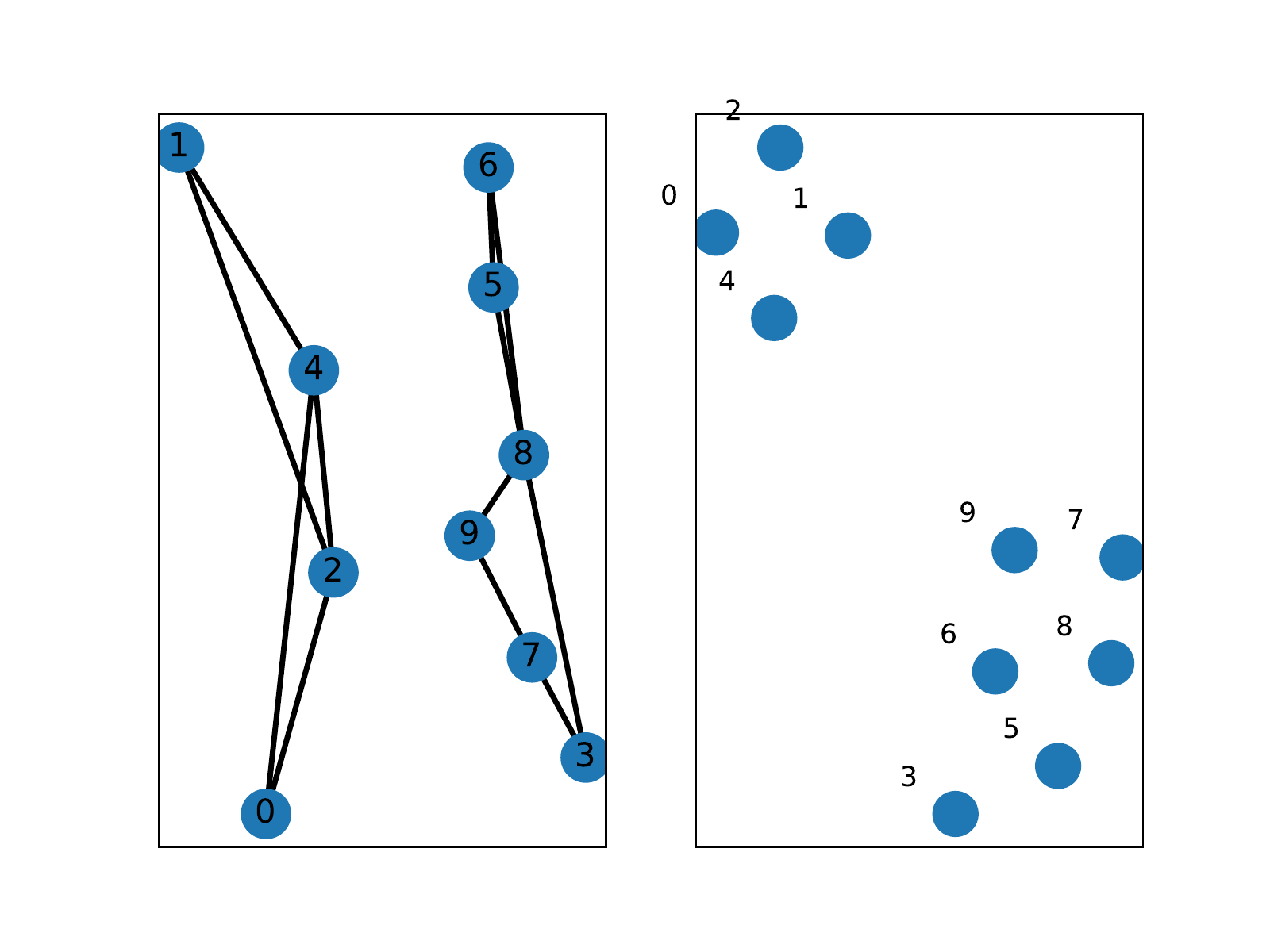}
\end{minipage}}  

\subfigure[12m\_vs\_14m(a)]{
\begin{minipage}[t]{0.2\textwidth}
\centering
\includegraphics[width=1.3in]{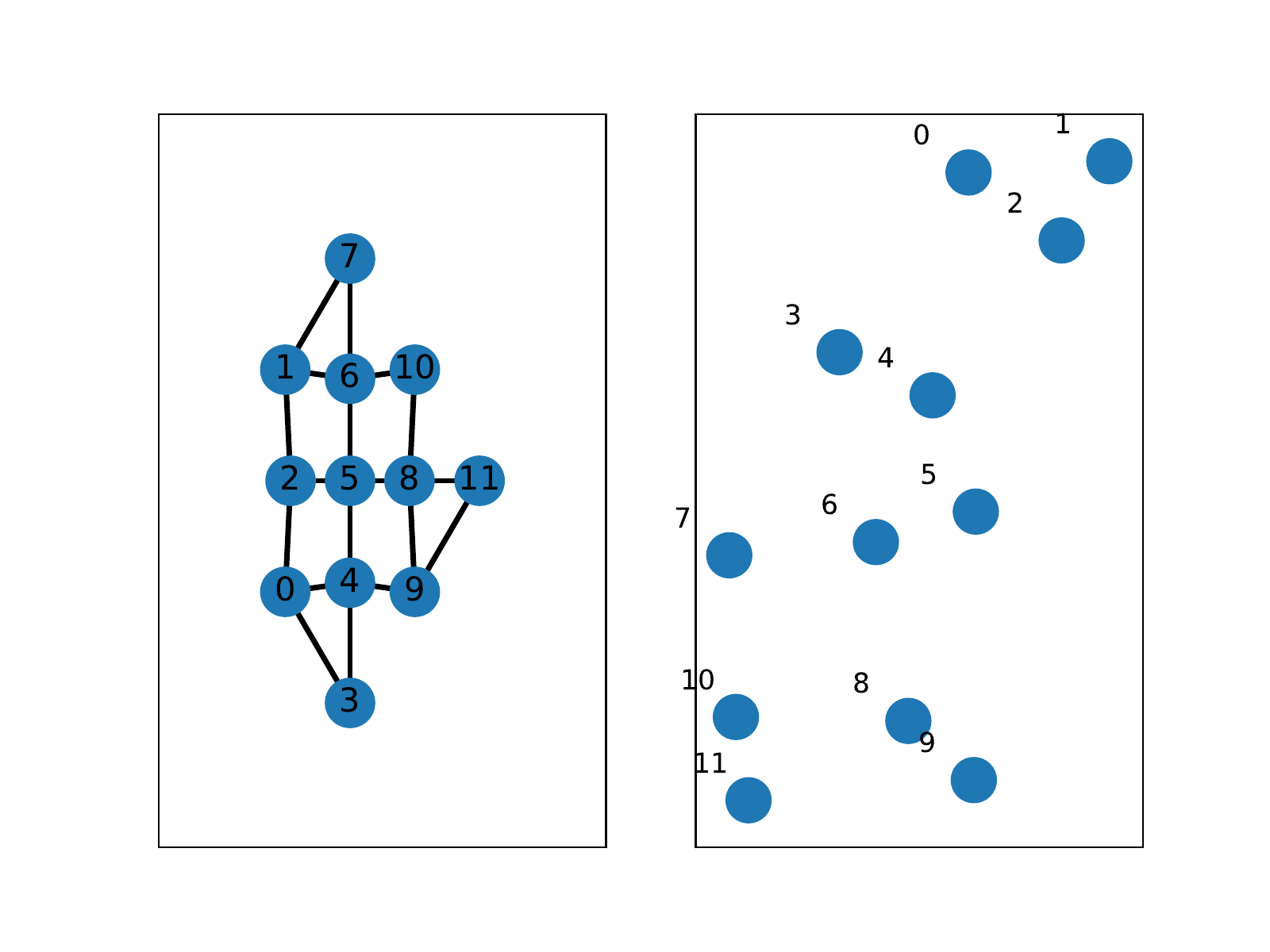}
\end{minipage}}
\subfigure[12m\_vs\_14m(b)]{
\begin{minipage}[t]{0.2\textwidth}
\centering
\includegraphics[width=1.3in]{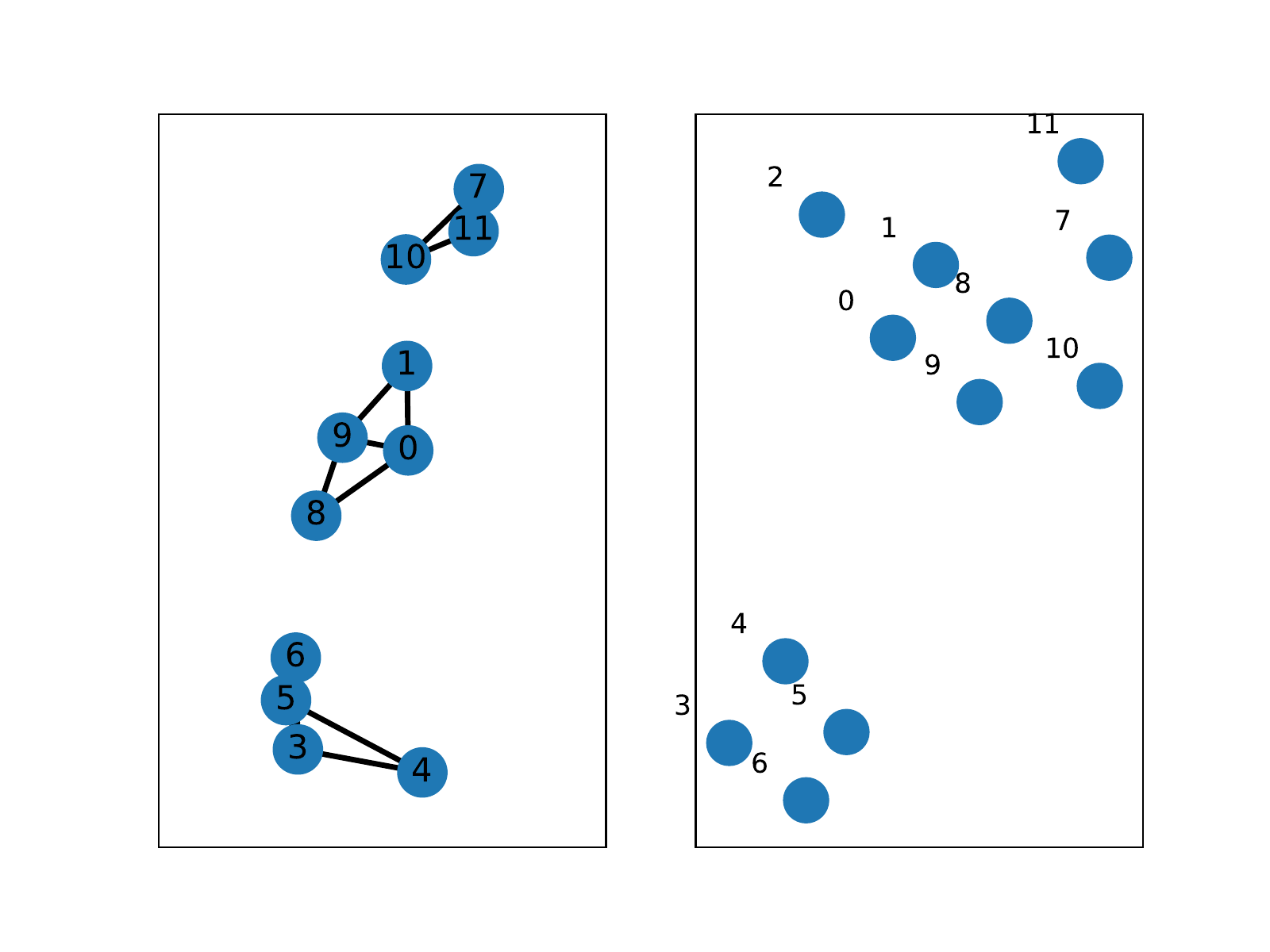}
\end{minipage}}  

\subfigure[15m\_vs\_17m(a)]{
\begin{minipage}[t]{0.2\textwidth}
\centering
\includegraphics[width=1.3in]{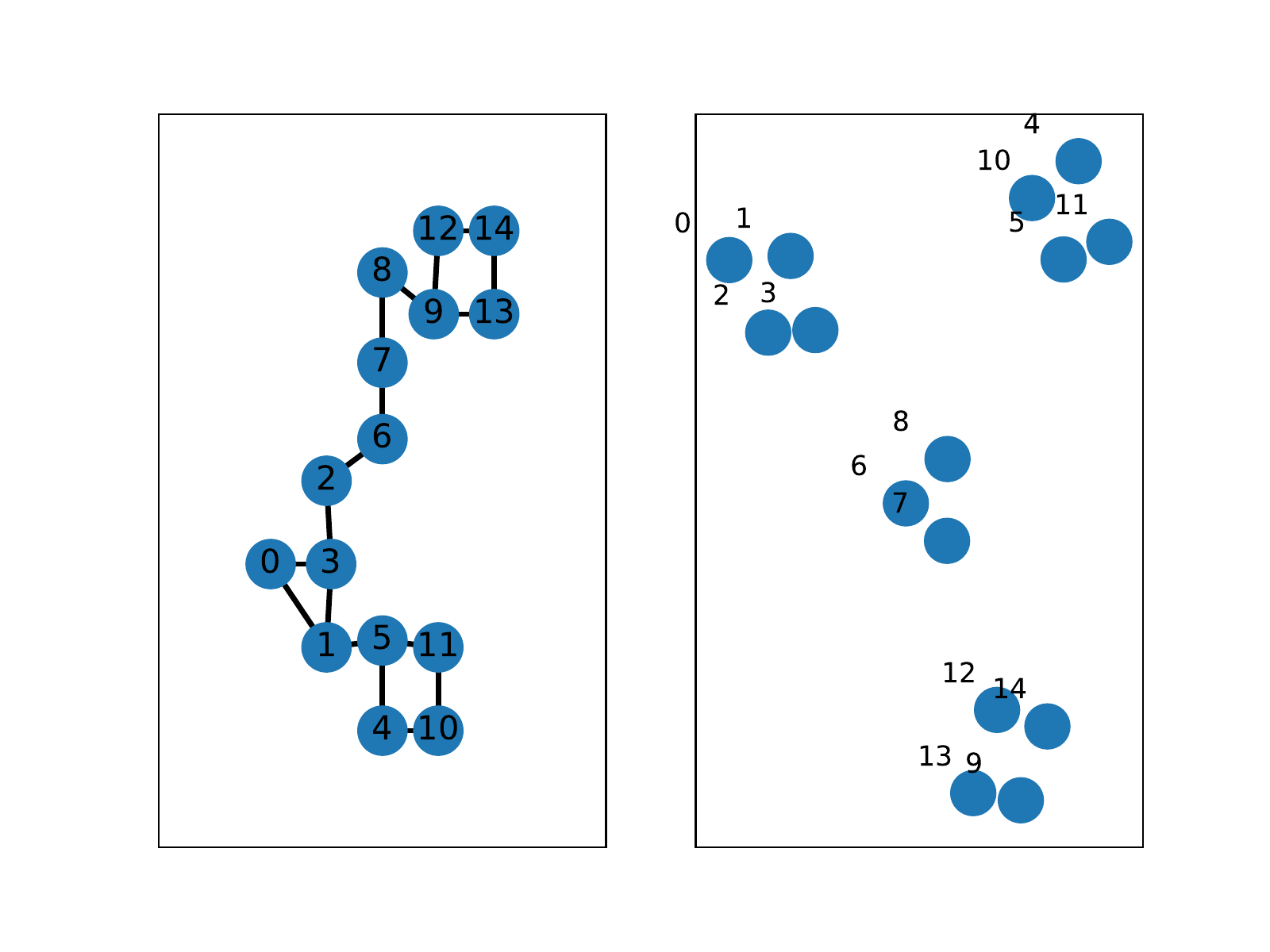}
\end{minipage}}
\subfigure[15m\_vs\_17m(b)]{
\begin{minipage}[t]{0.2\textwidth}
\centering
\includegraphics[width=1.3in]{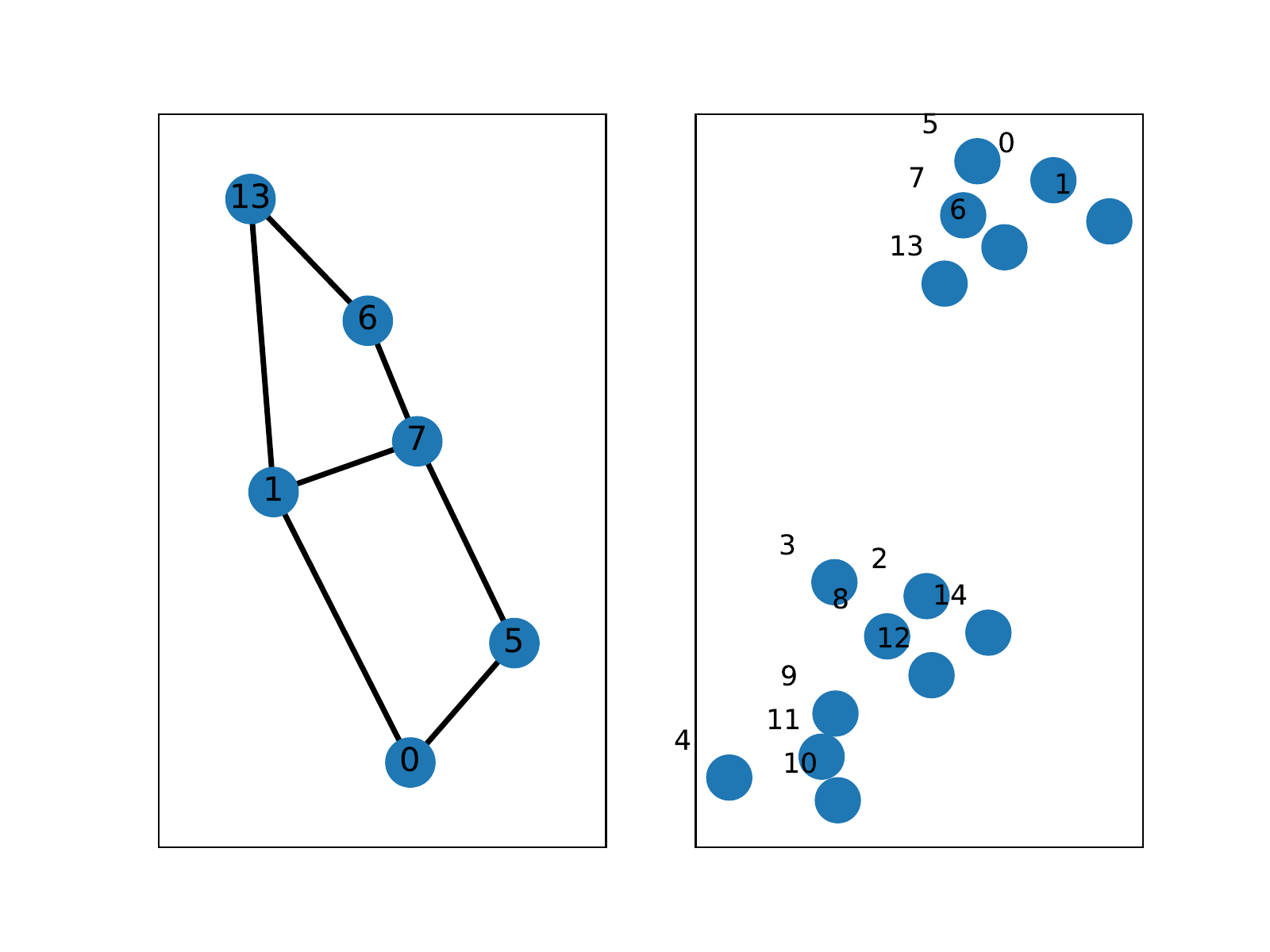}
\end{minipage}}  
            
\centering
\caption{Group feature representation base on t-SNE. }
\label{fig5}
\end{figure}

\subsection{Representation}
To demonstrate the superiority of our proposed algorithm, the t-SNE algorithm is used to reduce the group features' dimension for visually illustrating the effect of agent grouping. Pictures show the representation after compressing the group feature of the agent to two dimensions in multiple map scenarios. Pictures include various scenarios from the beginning of the game, the middle of the game, and the agent's death in the later stage. Figure \ref{fig5}(c) shows that BGC-QMIX first divides the agent group features into three groups at the beginning. Agents' group features are divided into two groups without any information interaction between groups with the game running. Figure \ref{fig5}(h) indicates that group features of the dead agents can be divided into a specific group that demonstrates the rationality of the algorithm. Moreover, we find that the agent mapping becomes more straightforward and more representative as the number of agents increases.

\subsection{Ablation}
To check the robustness of our proposed algorithm, an ablation experiment is conducted. The VDN network is used instead of the QMIX network to aggregate the independent Q values of the agent. Experimental results in 8m\_vs\_9m map, 10m\_vs\_11m map, and 12m\_vs\_14m map are compared. In two scenarios, where the number of agents is small, the group clustering-based method has a certain degree of improvement compared with the VDN and even the QMIX algorithm. In the more significant number 12m\_vs\_14m scenario, the VDN algorithm is better than that of QMIX; the VDN method based on graph clustering is still better than these two.

\subsection{Distributed Execution}
The introduced algorithm is tested within the distributed framework. The KL divergence method is used for the agents trained based on the CTCE framework to minimize the new group feature and the group feature after graph clustering. The overall algorithm training is about 0.5M steps, and the original network error on the test\_battle\_win\_mean index is within $\pm1 \%$.


\section{Conclusion}
This paper introduces the group concept into multi-agent reinforcement learning to achieve excellent performance in the non-communication setup. We design a novel agent structure to realize this non-communication goal via an individual agent module to generate individual features and an agent belief module to produce group features. Apply a fusion module to fuse these features to generate agent actions. In the agent belief module, we take the GAT network to merge the adjacent agent features and use a split loss to prevent all agents from being consistent.

Empirical results show that the proposed algorithm's effect is vastly improved compared to the current baseline algorithms, and this improvement increases with the increase in the number of agents. The t-SNE method is applied to verify that the group features of adjacent agents are more similar, indicating the agent group features representability under the graph attention network and the split loss. Additionally, the overall algorithm is integrated into the VDN network, and results show the performance improved compared to the original VDN network, even the QMIX method.

In future work, we will consider apply intrinsic rewards to group agents more efficiently and feasibly.	

\section{Acknowledgments}
	The work has been supported by grant number: Z181100003218013 from Beijing Municipal Science\&Technology Commission.

\bibliographystyle{ACM-Reference-Format} 
\bibliography{references.bib}

\begin{thebibliography}{26}
\providecommand{\natexlab}[1]{#1}
\providecommand{\url}[1]{\texttt{#1}}
\providecommand{\urlprefix}{URL }
\expandafter\ifx\csname urlstyle\endcsname\relax
  \providecommand{\doi}[1]{doi:\discretionary{}{}{}#1}\else
  \providecommand{\doi}{doi:\discretionary{}{}{}\begingroup
  \urlstyle{rm}\Url}\fi

\bibitem[{Agarwal, Kumar, and Sycara(2019)}]{LTCB}
Agarwal, A.; Kumar, S.; and Sycara, K.~P. 2019.
\newblock Learning Transferable Cooperative Behavior in Multi-Agent Teams.
\newblock \emph{CoRR} abs/1906.01202.
\newblock \urlprefix\url{http://arxiv.org/abs/1906.01202}.

\bibitem[{Busoniu, Babuska, and De~Schutter(2008)}]{marl}
Busoniu, L.; Babuska, R.; and De~Schutter, B. 2008.
\newblock A comprehensive survey of multiagent reinforcement learning.
\newblock \emph{IEEE Transactions on Systems, Man, and Cybernetics, Part C
  (Applications and Reviews)} 38(2): 156--172.

\bibitem[{Chung et~al.(2014)Chung, Gulcehre, Cho, and Bengio}]{gru}
Chung, J.; Gulcehre, C.; Cho, K.; and Bengio, Y. 2014.
\newblock Empirical evaluation of gated recurrent neural networks on sequence
  modeling.
\newblock \emph{arXiv preprint arXiv:1412.3555} .

\bibitem[{Ding, Huang, and Lu(2020)}]{LIIC}
Ding, Z.; Huang, T.; and Lu, Z. 2020.
\newblock Learning Individually Inferred Communication for Multi-Agent
  Cooperation.
\newblock \emph{arXiv preprint arXiv:2006.06455} .

\bibitem[{Du et~al.(2019)Du, Han, Fang, Liu, Dai, and Tao}]{LIIR}
Du, Y.; Han, L.; Fang, M.; Liu, J.; Dai, T.; and Tao, D. 2019.
\newblock LIIR: Learning Individual Intrinsic Reward in Multi-Agent
  Reinforcement Learning.
\newblock In Wallach, H.; Larochelle, H.; Beygelzimer, A.; d\textquotesingle
  Alch\'{e}-Buc, F.; Fox, E.; and Garnett, R., eds., \emph{Advances in Neural
  Information Processing Systems 32}, 4403--4414. Curran Associates, Inc.
\newblock
  \urlprefix\url{http://papers.nips.cc/paper/8691-liir-learning-individual-intrinsic-reward-in-multi-agent-reinforcement-learning.pdf}.

\bibitem[{Dudani(1976)}]{knn}
Dudani, S.~A. 1976.
\newblock The distance-weighted k-nearest-neighbor rule.
\newblock \emph{IEEE Transactions on Systems, Man, and Cybernetics} (4):
  325--327.

\bibitem[{Foerster et~al.(2016)Foerster, Assael, de~Freitas, and
  Whiteson}]{RIALDIAL}
Foerster, J.~N.; Assael, Y.~M.; de~Freitas, N.; and Whiteson, S. 2016.
\newblock Learning to Communicate with Deep Multi-Agent Reinforcement Learning.
\newblock \emph{CoRR} abs/1605.06676.
\newblock \urlprefix\url{http://arxiv.org/abs/1605.06676}.

\bibitem[{Ha, Dai, and Le(2016)}]{hypernetwork}
Ha, D.; Dai, A.; and Le, Q.~V. 2016.
\newblock Hypernetworks.
\newblock \emph{arXiv preprint arXiv:1609.09106} .

\bibitem[{Hausknecht and Stone(2015)}]{deep}
Hausknecht, M.; and Stone, P. 2015.
\newblock Deep recurrent q-learning for partially observable mdps.
\newblock In \emph{2015 AAAI Fall Symposium Series}.

\bibitem[{H{\"{u}}ttenrauch, Sosic, and Neumann(2017)}]{abs-1709-06011}
H{\"{u}}ttenrauch, M.; Sosic, A.; and Neumann, G. 2017.
\newblock Guided Deep Reinforcement Learning for Swarm Systems.
\newblock \emph{CoRR} abs/1709.06011.
\newblock \urlprefix\url{http://arxiv.org/abs/1709.06011}.

\bibitem[{Jiang, Dun, and Lu(2018)}]{GCRL}
Jiang, J.; Dun, C.; and Lu, Z. 2018.
\newblock Graph Convolutional Reinforcement Learning for Multi-Agent
  Cooperation.
\newblock \emph{CoRR} abs/1810.09202.
\newblock \urlprefix\url{http://arxiv.org/abs/1810.09202}.

\bibitem[{Kraemer and Banerjee(2016)}]{ctde2}
Kraemer, L.; and Banerjee, B. 2016.
\newblock Multi-agent reinforcement learning as a rehearsal for decentralized
  planning.
\newblock \emph{Neurocomputing} 190(may 19): 82--94.

\bibitem[{Kusner, Paige, and Hern{\'a}ndez-Lobato(2017)}]{vae}
Kusner, M.~J.; Paige, B.; and Hern{\'a}ndez-Lobato, J.~M. 2017.
\newblock Grammar variational autoencoder.
\newblock \emph{arXiv preprint arXiv:1703.01925} .

\bibitem[{Laurens and Hinton(2008)}]{T-SNE}
Laurens, V. D.~M.; and Hinton, G. 2008.
\newblock Visualizing Data using t-SNE.
\newblock \emph{Journal of Machine Learning Research} 9(2605): 2579--2605.

\bibitem[{Liu et~al.(2020)Liu, Wang, Hu, Hao, Chen, and Gao}]{MAGA}
Liu, Y.; Wang, W.; Hu, Y.; Hao, J.; Chen, X.; and Gao, Y. 2020.
\newblock Multi-Agent Game Abstraction via Graph Attention Neural Network.
\newblock In \emph{AAAI}, 7211--7218.

\bibitem[{Mao et~al.(2019)Mao, Liu, Hao, Luo, Li, Zhang, Wang, and Xiao}]{NCC}
Mao, H.; Liu, W.; Hao, J.; Luo, J.; Li, D.; Zhang, Z.; Wang, J.; and Xiao, Z.
  2019.
\newblock Neighborhood Cognition Consistent Multi-Agent Reinforcement Learning.
\newblock \emph{arXiv preprint arXiv:1912.01160} .

\bibitem[{Oliehoek, Amato et~al.(2016)}]{ctde1}
Oliehoek, F.~A.; Amato, C.; et~al. 2016.
\newblock \emph{A concise introduction to decentralized POMDPs}, volume~1.
\newblock Springer.

\bibitem[{Peng et~al.(2017)Peng, Yuan, Wen, Yang, Tang, Long, and
  Wang}]{BICNET}
Peng, P.; Yuan, Q.; Wen, Y.; Yang, Y.; Tang, Z.; Long, H.; and Wang, J. 2017.
\newblock Multiagent Bidirectionally-Coordinated Nets for Learning to Play
  StarCraft Combat Games.
\newblock \emph{CoRR} abs/1703.10069.
\newblock \urlprefix\url{http://arxiv.org/abs/1703.10069}.

\bibitem[{Rashid et~al.(2018)Rashid, Samvelyan, de~Witt, Farquhar, Foerster,
  and Whiteson}]{QMIX}
Rashid, T.; Samvelyan, M.; de~Witt, C.~S.; Farquhar, G.; Foerster, J.~N.; and
  Whiteson, S. 2018.
\newblock {QMIX:} Monotonic Value Function Factorisation for Deep Multi-Agent
  Reinforcement Learning.
\newblock \emph{CoRR} abs/1803.11485.
\newblock \urlprefix\url{http://arxiv.org/abs/1803.11485}.

\bibitem[{Son et~al.(2019)Son, Kim, Kang, Hostallero, and Yi}]{QTRAN}
Son, K.; Kim, D.; Kang, W.~J.; Hostallero, D.; and Yi, Y. 2019.
\newblock {QTRAN:} Learning to Factorize with Transformation for Cooperative
  Multi-Agent Reinforcement Learning.
\newblock \emph{CoRR} abs/1905.05408.
\newblock \urlprefix\url{http://arxiv.org/abs/1905.05408}.

\bibitem[{Sunehag et~al.(2017)Sunehag, Lever, Gruslys, Czarnecki, Zambaldi,
  Jaderberg, Lanctot, Sonnerat, Leibo, Tuyls, and Graepel}]{VDN}
Sunehag, P.; Lever, G.; Gruslys, A.; Czarnecki, W.~M.; Zambaldi, V.~F.;
  Jaderberg, M.; Lanctot, M.; Sonnerat, N.; Leibo, J.~Z.; Tuyls, K.; and
  Graepel, T. 2017.
\newblock Value-Decomposition Networks For Cooperative Multi-Agent Learning.
\newblock \emph{CoRR} abs/1706.05296.
\newblock \urlprefix\url{http://arxiv.org/abs/1706.05296}.

\bibitem[{Van~Erven and Harremos(2014)}]{kl}
Van~Erven, T.; and Harremos, P. 2014.
\newblock R{\'e}nyi divergence and Kullback-Leibler divergence.
\newblock \emph{IEEE Transactions on Information Theory} 60(7): 3797--3820.

\bibitem[{Veli{\v{c}}kovi{\'c} et~al.(2017)Veli{\v{c}}kovi{\'c}, Cucurull,
  Casanova, Romero, Lio, and Bengio}]{gat}
Veli{\v{c}}kovi{\'c}, P.; Cucurull, G.; Casanova, A.; Romero, A.; Lio, P.; and
  Bengio, Y. 2017.
\newblock Graph attention networks.
\newblock \emph{arXiv preprint arXiv:1710.10903} .

\bibitem[{Vinyals et~al.(2017)Vinyals, Ewalds, Bartunov, Georgiev, Vezhnevets,
  Yeo, Makhzani, K{\"{u}}ttler, Agapiou, Schrittwieser, Quan, Gaffney,
  Petersen, Simonyan, Schaul, van Hasselt, Silver, Lillicrap, Calderone, Keet,
  Brunasso, Lawrence, Ekermo, Repp, and Tsing}]{SMAC}
Vinyals, O.; Ewalds, T.; Bartunov, S.; Georgiev, P.; Vezhnevets, A.~S.; Yeo,
  M.; Makhzani, A.; K{\"{u}}ttler, H.; Agapiou, J.; Schrittwieser, J.; Quan,
  J.; Gaffney, S.; Petersen, S.; Simonyan, K.; Schaul, T.; van Hasselt, H.;
  Silver, D.; Lillicrap, T.~P.; Calderone, K.; Keet, P.; Brunasso, A.;
  Lawrence, D.; Ekermo, A.; Repp, J.; and Tsing, R. 2017.
\newblock StarCraft {II:} {A} New Challenge for Reinforcement Learning.
\newblock \emph{CoRR} abs/1708.04782.
\newblock \urlprefix\url{http://arxiv.org/abs/1708.04782}.

\bibitem[{Wang et~al.(2019)Wang, Wang, Wu, and Zhang}]{IBM}
Wang, T.; Wang, J.; Wu, Y.; and Zhang, C. 2019.
\newblock Influence-based multi-agent exploration.
\newblock \emph{arXiv preprint arXiv:1910.05512} .

\bibitem[{Wang et~al.(2020)Wang, Yang, Liu, Hao, Hao, Hu, Chen, Fan, and
  Gao}]{FFTM}
Wang, W.; Yang, T.; Liu, Y.; Hao, J.; Hao, X.; Hu, Y.; Chen, Y.; Fan, C.; and
  Gao, Y. 2020.
\newblock From Few to More: Large-Scale Dynamic Multiagent Curriculum Learning.
\newblock In \emph{AAAI}, 7293--7300.

\end{thebibliography}
\end{document}


\section{appendix}

\subsection{QMIX Architecture}
\begin{figure}[htbp]
\centering
\includegraphics[width=0.8\columnwidth]{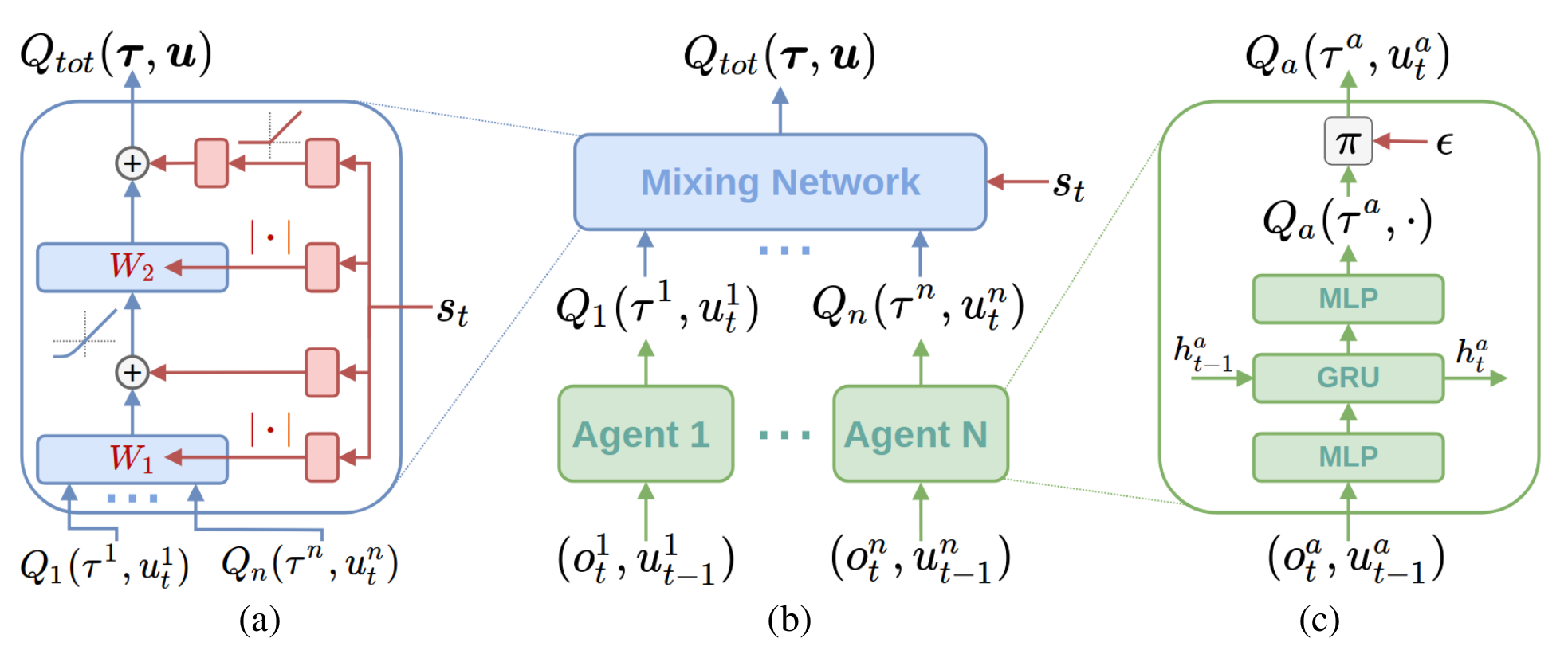}
\caption{The schematics of QMIX. (a) The structure of Mixing Network. This module uses hypernetwork to merge all individual Q value with the monotonic constraint. (b) The overall of QMIX architecture. (c) The structure of agent network with GRU cell.}
\label{fig1}
\end{figure}

\subsection{Experimental Setup}

\subsubsection{Architecture and Training}
\begin{figure}[htbp]
\centering
\includegraphics[width=0.8\columnwidth]{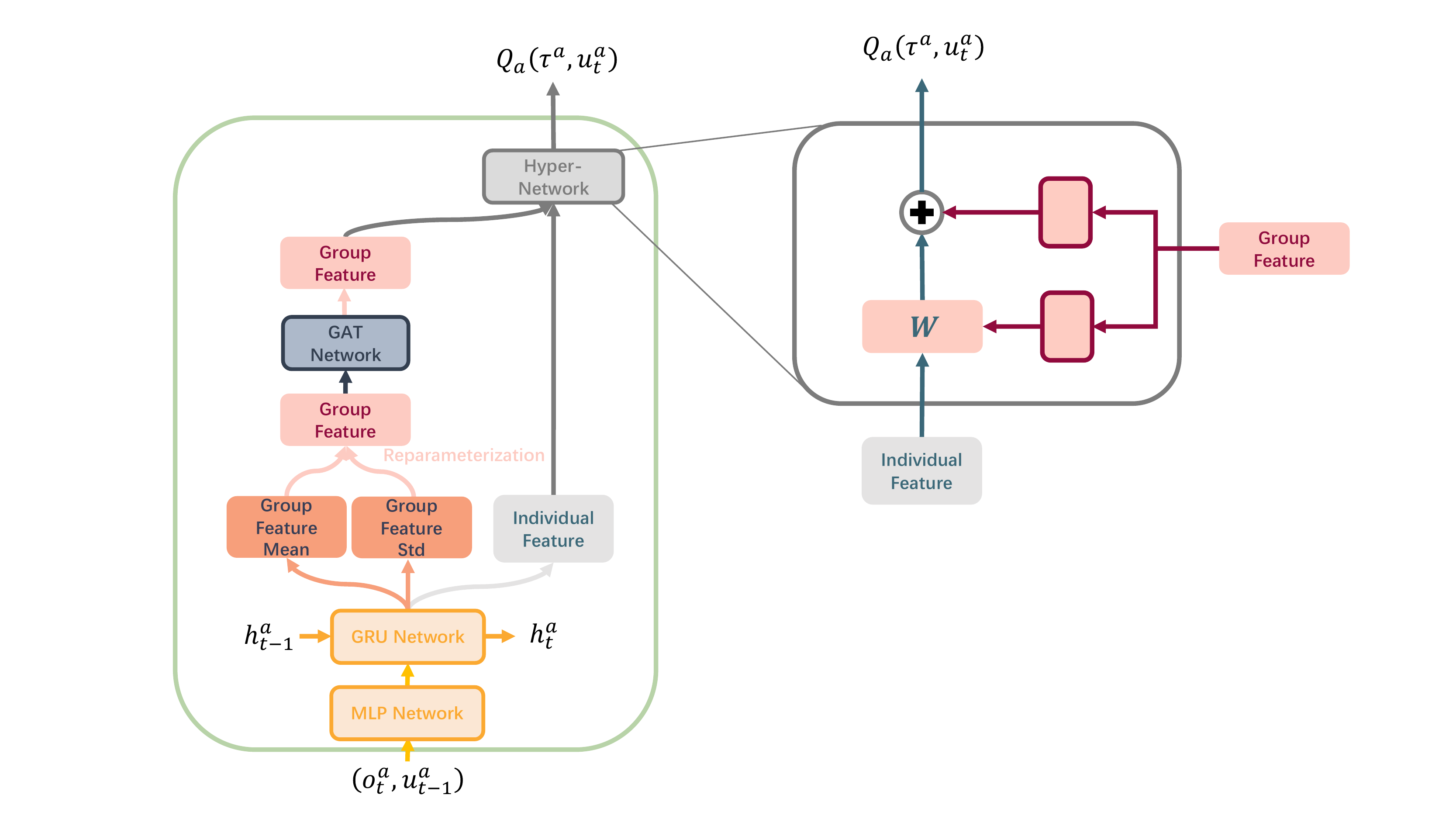}
\caption{ The Struct of Agent.}
\label{fig2}
\end{figure}

Figure \ref{fig2} shows the agent's structure. At each step, each agent receives its local observation and transfers it into the agent group feature and individual feature. As shown in Figure \ref{fig2}, the group feature is feed into two different networks to generate two new features. These two features are the weight and bias of the fully connected network which generates the individual Q value.

Since each agent shares the same network, we add the agent's id information via one-hot encoding into the observation input to distinguish the agent. In addition, we concatenate agent's last action to the observation.

\begin{table}[htbp]
\caption{Algorithm Parameters}
\label{tab:1}      
\begin{tabular}{lll}
\hline\noalign{\smallskip}
Parameter & Value \\
\noalign{\smallskip}\hline\noalign{\smallskip}
RNN Hidden Dim &  64 \\
Group Feature Dim &  32 \\
Individual Feature Dim &  32 \\
GAT Dropout & 0.5 \\
GAT LeakyRelu Alpha & 0.2 \\
KL Constraint \delta & 0.005 \\
QMIX Hypernet Dim & 64 \\
QMIX Layer & 2 \\
\noalign{\smallskip}\hline
\end{tabular}
\end{table}

\subsubsection{SMAC Environment}

To enhance the exploration ability of QMIX Baseline, the agent's strategy adopts the $\epsilon$-greedy strategy, $\epsilon$ anneals from 1.0 to 0.05 with anneal time 50k. 
In addition, to reduce the algorithm training time, we train the agent in parallel. We set the batch\_size\_run parameter to 7 due to the CPU limitation.

\begin{table}[h]
\caption{SMAC Parameters}
\label{tab:2}       
\begin{tabular}{lll}
\hline\noalign{\smallskip}
Parameter & Value \\
\noalign{\smallskip}\hline\noalign{\smallskip}
Buffer Size &  5000 \\
Final Epsilon Value & 0.05 \\
Batch Size & 7 \\
\noalign{\smallskip}\hline
\end{tabular}
\end{table}